\pdfoutput=1
\documentclass[11pt]{article}
\usepackage[final]{acl}
\usepackage{times}
\usepackage{latexsym}
\usepackage[T1]{fontenc}
\usepackage[utf8]{inputenc}
\usepackage{microtype}
\usepackage{inconsolata}
\usepackage{adjustbox}
\usepackage{amsmath}
\usepackage{amsthm}
\usepackage{amssymb}
\usepackage{array}
\usepackage{booktabs}
\usepackage{colortbl}
\usepackage{enumitem}
\usepackage{float}
\usepackage{graphicx}
\usepackage{multirow}
\usepackage{pifont}
\usepackage{relsize}
\usepackage{tabularx}
\usepackage[most]{tcolorbox}
\usepackage{url}
\usepackage{xcolor}
\usepackage{xspace}

\setkeys{Gin}{draft=false}
\newcommand{\fsYes}{\ensuremath{\checkmark}}
\newcommand{\fsPart}{\ensuremath{\circ}}
\newcommand{\fsNo}{\ensuremath{\times}}

\definecolor{lightyellow}{RGB}{255,249,230}

\definecolor{yellow}{RGB}{255, 255, 150}
\definecolor{lightblue}{RGB}{173, 216, 230}
\definecolor{lightred}{RGB}{255, 182, 193}
\definecolor{lightgreen}{RGB}{144, 238, 144}

\definecolor{greenbox}{RGB}{144, 238, 144}
\definecolor{redbox}{RGB}{255, 182, 193}
\definecolor{bluebox}{RGB}{135, 206, 235}
\definecolor{yellowbox}{RGB}{255, 255, 0}
\definecolor{posgreen}{RGB}{198, 239, 206}
\definecolor{negred}{RGB}{255, 199, 206}

\tcbset{
    eval_prompt/.style={
        colback=gray!5,
        colframe=black!70,
        boxrule=0.5pt,
        arc=2mm,
        left=8pt, right=8pt, top=8pt, bottom=8pt,
        fontupper=\small,
        width=\linewidth
    }
}

\newtcolorbox{promptbox}[1][]{%
  enhanced,
  colback=black!3,
  colframe=black!75,
  boxrule=0.7pt,
  arc=5pt,
  left=8pt,
  right=8pt,
  top=7pt,
  bottom=7pt,
  boxsep=0pt,
  width=\linewidth,
  fontupper=\footnotesize,
  before skip=0pt,
  after skip=0pt,
  #1
}

\newcommand{\logo}{\raisebox{-0.32em}{%
  \includegraphics[height=2.5em]{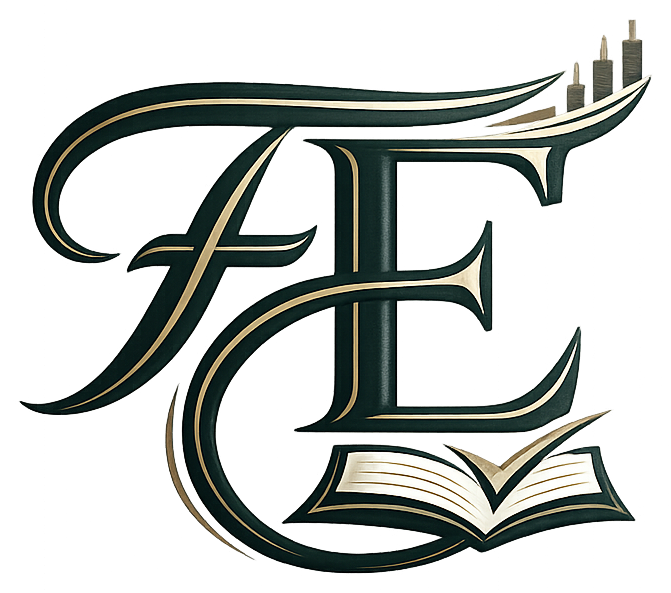}}\hspace{0em}}

\title{\logo \textsc{FinExam-10K}: When Retrieval Helps Financial Reasoning?}

\author{
\textbf{Yan Lin}\textsuperscript{1,2},
\textbf{Jingyu Sun}\textsuperscript{3,4},
\textbf{Zhongliang Guo}\textsuperscript{5},\\
\textbf{Qing Li}\textsuperscript{6},
\textbf{Zhuohan Xie}\textsuperscript{7},
\textbf{Yuxia Wang}\textsuperscript{1,\textdagger} \\
\vspace{0.5mm}
\textsuperscript{1}INSAIT, Sofia University ``St. Kliment Ohridski''\quad
\textsuperscript{2}Newcastle University\quad \\
\textsuperscript{3}University of Manchester\quad
\textsuperscript{4}University of Melbourne\quad
\textsuperscript{5}University of Aberdeen\quad \\
\textsuperscript{6}University of Groningen\quad
\textsuperscript{7}MBZUAI \\
\textsuperscript{\textdagger}Corresponding author \\
\vspace{1mm}
\texttt{y.lin64@ncl.ac.uk, yuxia.wang@insait.ai}
}

\begin{document}
\maketitle
\begin{abstract} 
Professional financial examinations require models to combine domain knowledge, calculation, and judgment, yet no benchmark covers the full CFA and FRM structure under one protocol. We introduce \textsc{FinExam-10K}, to our knowledge the largest reported English benchmark for this setting, with 10{,}198 expert-reannotated questions spanning CFA Levels I--III and FRM Parts I--II. We release 5{,}110 questions and sequester 5{,}088 for a quarterly maintained leaderboard. To separate coverage from local answerability, we report a 10{,}198-item \emph{Full-Coverage Track} and a 7{,}625-item \emph{Context-Complete Reasoning Track}, which is the primary basis for claims about reasoning from the supplied record. Across 17 models, the best accuracy is 85.29\% overall. On the frozen Hard band, the best score is 34.68\% on the Full-Coverage Track and 54.57\% on the 372 context-complete items. All 17 models share 47 context-complete failures. Function-RAG and FunctionGraph-RAG rescue hundreds of errors but also overturn many correct answers, producing little or negative net gain. A gate trained only on public data decides from the question and initial response when FunctionGraph-RAG should run. On the 5{,}088 held-out items, the gate invokes FunctionGraph-RAG for 7.9\% of questions and improves accuracy from 70.83\% to 71.23\% (\(p=.0446\)). Dataset and code are
available at \href{https://anonymous.4open.science/r/FinExam10k-2A08/README.md}{FinExam-10K}.
\end{abstract}

\section{Introduction}

\begin{figure}[!t]
    \centering
    \includegraphics[width=0.56\textwidth]{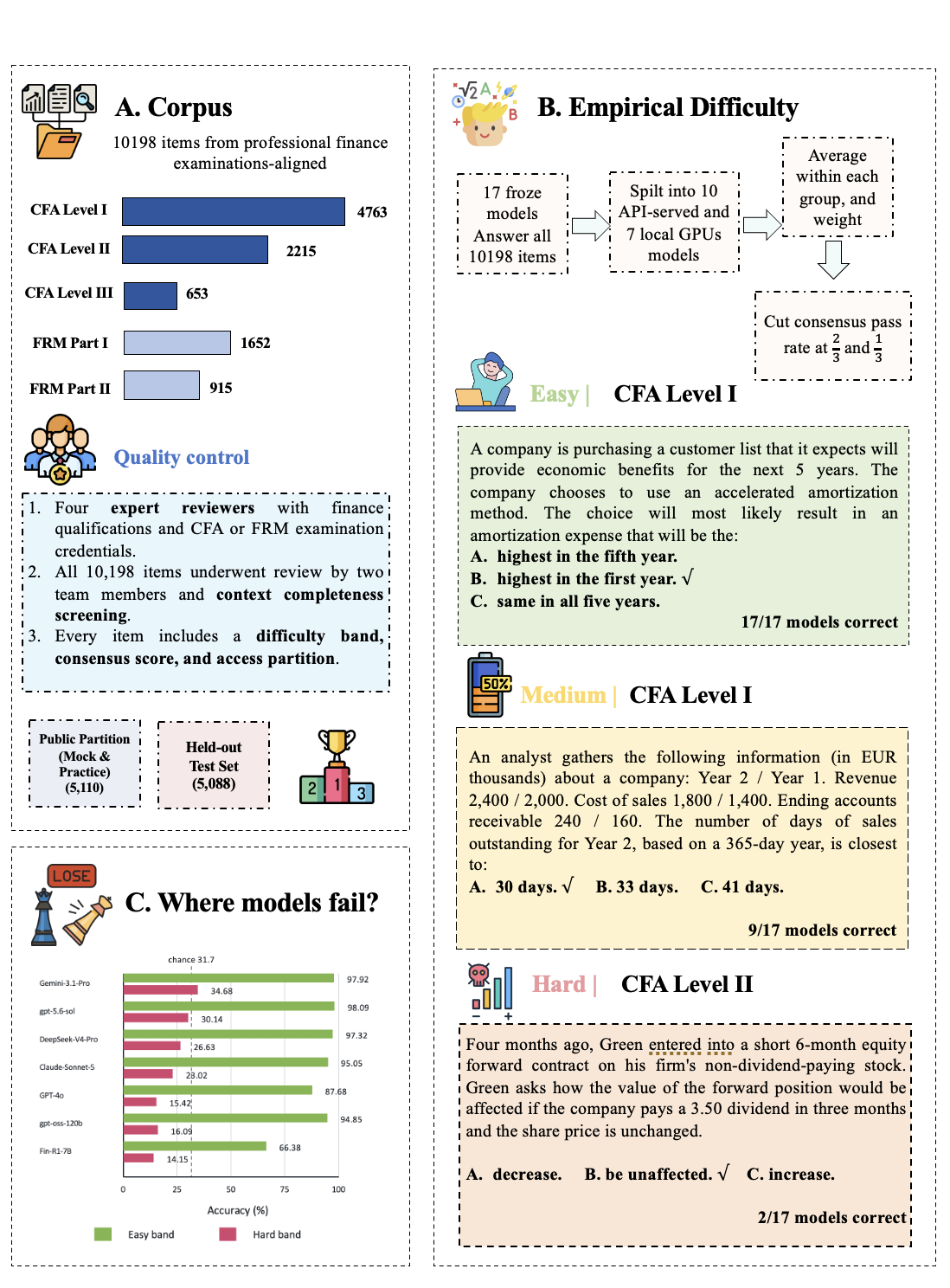}
    \caption{\textbf{FinExam-10K benchmark design and empirical difficulty.} (A) The 10{,}198 questions span five CFA and FRM stages. (B) Consensus defines Easy, Medium, and Hard bands, illustrated by one representative item from each band. (C) Across representative models, accuracy falls sharply from Easy to Hard.}
    \label{fig:overview}
\end{figure}

Professional financial reasoning requires more than recalling definitions or substituting values into isolated formulas. The CFA Program and the FRM certification define externally specified curricula in investment management and financial risk. Across CFA Levels I--III and FRM Parts I--II, questions increasingly combine domain knowledge, quantitative operations, rule selection, temporal or contractual constraints, and professional judgment. These five stages therefore provide an interpretable setting for locating model failures against standards defined independently of the benchmark authors.

Existing financial benchmarks cover report-grounded question answering~\citep{chen-etal-2021-finqa}, financial mathematics~\citep{zhao-etal-2024-knowledgefmath}, selected professional assessments~\citep{mahfouz-etal-2024-state, yao-etal-2025-evaluating}, and broad certification collections~\citep{guo-etal-2025-flame,zhang-etal-2026-fire}. However, they do not provide a unified, level-resolved benchmark spanning both CFA and FRM with item-aligned intervention analysis under one protocol. This gap prevents controlled comparisons of where a model fails, whether external knowledge repairs that item, and whether the same intervention damages an answer that was already correct.

We introduce \textsc{FinExam-10K}, a benchmark of 10{,}198 English multiple-choice questions covering all five stages. A four-member finance-qualified team conducted stage-matched primary review and a complete second pass over every retained item. We release 5{,}110 questions identified as mock or practice material and reserve 5{,}088 for held-out evaluation. To the best of our knowledge, \textsc{FinExam-10K} is the largest reported English benchmark dedicated to CFA- and FRM-aligned examination reasoning and the first to preserve all five stages under one protocol. We further establish a leaderboard and a versioned quarterly refresh.

Every record contains a stem, answer options, a gold answer, and a reference rationale. Context completeness is a separate representation-level property. Professional examinations often place several subquestions under a shared vignette, table, image, or exhibit; flattening such item sets can detach an otherwise complete record from evidence introduced elsewhere in the source set. To separate curriculum coverage from reasoning grounded in the supplied record, we define two complementary tracks. The 10{,}198-item \emph{Full-Coverage Track} preserves the complete curated item universe and supports the maintained leaderboard. The 7{,}625-item \emph{Context-Complete Reasoning Track} retains only questions whose answer-necessary evidence is locally available and is the primary basis for claims about reasoning from the supplied record. Both tracks preserve the original public/held-out partition and frozen difficulty labels.

A fixed panel of 17 models defines an empirical difficulty axis rather than inheriting difficulty from curriculum stage. The resulting partition contains 6{,}578 Easy, 2{,}183 Medium, and 1{,}437 Hard items. The best overall model reaches 85.29\%, yet the highest Hard-band accuracy is only 34.68\%. Moreover, all 17 models fail on the same 188 items, and errors on context-complete Hard questions concentrate on the same distractors far more often than an item-specific random-choice null predicts. These analyses expose shared challenge patterns that aggregate accuracy and examination stage alone do not reveal.

We then study whether structured financial knowledge repairs these failures. Following FinanceReasoning~\citep{tang-etal-2025-financereasoning}, we evaluate matched Function-RAG baselines and two chain-specific FunctionGraph-RAG selectors trained on FinanceReasoning relevance labels and transferred without adaptation under DeepSeek-R1 CoT and GPT-4o PoT. Static augmentation rescues hundreds of Direct errors, but its gains are offset by intervention-induced harms. We therefore train a Direct-conditioned gate using only the 5{,}110 public items. The gate observes the question and signals available after the Direct call, then decides whether FunctionGraph-RAG should be invoked. On the 5{,}088 held-out items, it triggers 404 times and raises accuracy from 70.83\% to 71.23\% ($p=.0446$). This establishes a bounded form of adaptive control without claiming a full agentic reasoning system.

Our contributions are threefold:
\begin{itemize}[leftmargin=*,nosep]
    \item We introduce the largest reported English CFA- and FRM-aligned examination reasoning benchmark, with 10{,}198 expert-reannotated questions, public and held-out partitions, a Full-Coverage Track, a Context-Complete Reasoning Track, a maintained leaderboard, and quarterly versioning.
    \item We establish a frozen 17-model leaderboard and consensus-based Easy, Medium, and Hard bands, separating Full-Coverage degradation from context-complete Hard reasoning and identifying 47 universal failures with complete local evidence.
    \item We conduct matched Function-RAG, FunctionGraph-RAG, verification, and held-out gating experiments, showing that static retrieval has conditional value but requires selective invocation to avoid disrupting correct reasoning.
\end{itemize}

\section{Related Work}
\label{sec:relatedwork}

\begin{table*}[t]
\centering
\setlength{\abovecaptionskip}{4pt}
\setlength{\belowcaptionskip}{0pt}
\scriptsize
\setlength{\tabcolsep}{2.4pt}
\renewcommand{\arraystretch}{1.1}

\begin{adjustbox}{max width=\textwidth,center}
\begin{tabular}{@{}lll*{9}{c}@{}}
\toprule
\textbf{Benchmark}
& \textbf{Primary Scope}
& \textbf{\#Size}
& \textbf{C3}
& \textbf{F2}
& \textbf{ER}
& \textbf{RR}
& \textbf{DG}
& \textbf{OP}
& \textbf{SQ}
& \textbf{LB}
& \textbf{IT}
\\
\midrule

\multicolumn{12}{@{}l}{
\textit{Financial QA and Reasoning Benchmarks}
}\\[-1pt]

FinQA~\citep{chen-etal-2021-finqa}
& Report-based numerical QA
& 8,281
& \fsNo
& \fsNo
& \fsYes
& \fsYes
& \fsNo
& \fsYes
& \fsNo
& \fsNo
& \fsNo
\\

FinanceMATH~\citep{zhao-etal-2024-knowledgefmath}
& Financial math reasoning
& 1,200
& \fsNo
& \fsNo
& \fsYes
& \fsYes
& \fsNo
& \fsYes
& \fsNo
& \fsYes
& \fsPart
\\

XFinBench~\citep{zhang-etal-2025-xfinbench}
& Graduate-level finance
& 4,235
& \fsNo
& \fsNo
& \fsPart
& \fsPart
& \fsNo
& \fsYes
& \fsNo
& \fsNo
& \fsPart
\\

FinanceReasoning~\citep{tang-etal-2025-financereasoning}
& Financial numerical reasoning
& 2,238
& \fsNo
& \fsNo
& \fsYes
& \fsYes
& \fsYes
& \fsYes
& \fsNo
& \fsNo
& \fsPart
\\

\midrule

\multicolumn{12}{@{}l}{
\textit{Professional Financial Examination Benchmarks}
}\\[-1pt]

CFA Exams~\citep{mahfouz-etal-2024-state}
& CFA Levels I--III
& 1,164
& \fsYes
& \fsNo
& \fsPart
& \fsNo
& \fsNo
& \fsNo
& \fsNo
& \fsNo
& \fsPart
\\

CFA-RAG Study~\citep{yao-etal-2025-evaluating}
& CFA Levels I--III with RAG
& 1,560
& \fsYes
& \fsNo
& \fsPart
& \fsNo
& \fsNo
& \fsNo
& \fsNo
& \fsNo
& \fsPart
\\

RealFin~\citep{dai-etal-2026-realfin}
& CFA and CPA missing-premise reasoning
& 367 EN + 175 ZH
& \fsNo
& \fsNo
& \fsNo
& \fsNo
& \fsNo
& \fsYes
& \fsNo
& \fsNo
& \fsNo
\\

FLAME-Cer~\citep{guo-etal-2025-flame}
& 14 financial certifications
& $\sim$16K mixed
& \fsPart
& \fsPart
& \fsYes
& \fsNo
& \fsNo
& \fsPart
& \fsNo
& \fsPart
& \fsNo
\\

FIRE~\citep{zhang-etal-2026-fire}
& Certifications and scenarios
& 14K + 3K mixed
& \fsPart
& \fsPart
& \fsPart
& \fsPart
& \fsNo
& \fsYes
& \fsNo
& \fsNo
& \fsNo
\\

FINESSE-Bench~\citep{stanishevskii-etal-2026-finesse}
& CFA-like and trading tasks
& 3,993 mixed
& \fsPart
& \fsNo
& \fsPart
& \fsNo
& \fsNo
& \fsYes
& \fsNo
& \fsNo
& \fsNo
\\

\midrule

\rowcolor[gray]{0.92}
\textbf{FinExam-10K (Ours)}
& \textbf{English CFA/FRM-aligned exam reasoning}
& \textbf{10,198}
& \fsYes
& \fsYes
& \fsYes
& \fsYes
& \fsYes
& \fsYes
& \fsYes
& \fsYes
& \fsYes
\\

\bottomrule
\end{tabular}
\end{adjustbox}
\caption{
Comparison of FinExam-10K with representative financial reasoning and professional examination benchmarks.
\fsYes{} denotes full support, \fsPart{} partial or coarse support, and
\fsNo{} no support.
\textbf{C3} and \textbf{F2} indicate separate preservation of CFA
Levels I--III and FRM Parts I--II.
\textbf{ER} denotes full-item expert re-annotation and audit,
\textbf{RR} item-level reference rationales or executable solutions,
and \textbf{DG} explicit Easy, Medium, and Hard difficulty labels.
\textbf{OP} denotes a public item-level split, \textbf{SQ} a
sequestered evaluation set, and \textbf{LB} a public leaderboard.
\textbf{IT} denotes item-aligned analysis of rescued errors,
intervention-induced errors, and persistent failures.
Counts marked as mixed combine multiple certifications, languages,
or task families.
}
\label{tab:benchmark-comparison}
\end{table*}
\vspace{-0.45\baselineskip}

\paragraph{Financial Models and Reasoning Benchmarks.}
Financial NLP has progressed from domain-adapted models such as BloombergGPT\citep{wu-etal-2023-bloomberggpt}, FinGPT\citep{yang-etal-2023-fingpt}, and PIXIU\citep{xie-etal-2023-pixiu}.
Recent finance-specialized models include Fino1\citep{qian-etal-2025-fino1}, Fin-R1\citep{liu-etal-2025-finr1}, DianJin-R1\citep{zhu-etal-2025-dianjinr1}, Hawkish-8B\citep{mukaj-2024-hawkish} and ODA-Fin-RL\citep{cao-etal-2026-unlocking}. 
Evaluation has evolved in parallel, from report-grounded numerical QA in FinQA\citep{chen-etal-2021-finqa}, TAT-QA\citep{zhu-etal-2021-tat}, and ConvFinQA\citep{chen-etal-2022-convfinqa} to knowledge-intensive and executable reasoning in FinanceMATH\citep{zhao-etal-2024-knowledgefmath}, XFinBench\citep{zhang-etal-2025-xfinbench}, FinanceReasoning\citep{tang-etal-2025-financereasoning}, FinChain\citep{xie-etal-2026-finchain}, and AlphaQT-Bench\citep{luo-etal-2026-alphaqt}. 
These resources expose important financial reasoning failures, but they do not organize evaluation around one complete multi-stage professional curriculum.

\paragraph{Professional Financial Examinations.} Professional examinations provide externally defined curricula and human reference standards. Existing English studies have concentrated on CFA, including evaluation across all three levels\citep{mahfouz-etal-2024-state}, deeper assessment of Level III\citep{shetty-etal-2025-advanced}, curriculum-grounded retrieval on 1,560 questions\citep{yao-etal-2025-evaluating}, and recent reasoning models on 980 questions\citep{patel-etal-2025-reasoning}.
FINESSE-Bench\citep{stanishevskii-etal-2026-finesse} extends evaluation to CFA-like and technical-analysis tasks, RealFin\citep{dai-etal-2026-realfin} examines underdetermination in English-CFA and Chinese-CPA, and FLAME-Cer\citep{guo-etal-2025-flame} and FIRE\citep{zhang-etal-2026-fire} broaden coverage across multipl certifications and languages.
As Table~\ref{tab:benchmark-comparison} shows, prior English resources are CFA-only, CFA-like, or broad suites with coarse certification labels. FinExam-10K instead preserves CFA Levels I--III and FRM Parts I--II as five distinct stages under one protocol.

\paragraph{Structured Knowledge Augmentation and Verification.} FinanceReasoning\citep{tang-etal-2025-financereasoning} is our closest methodological predecessor. Its results show that passage retrieval can reduce accuracy, whereas instructed function retrieval with relevance filtering gives the strongest results for GPT-4o with PoT and DeepSeek-R1 with CoT.
Recent work further structures financial evidence through document hierarchies\citep{zou-etal-2026-fin}, analyst cards\citep{zhou-etal-2026-fincards}, multimodal retrieval\citep{yang-etal-2026-finmragbench}, and large tool spaces\citep{huang-etal-2026-fintoolsyn}. Verification has likewise moved toward financial error detection\citep{he-etal-2026-large},
executable traces\citep{xie-etal-2026-finchain}, symbolic graph checks\citep{wang-etal-2026-sr}, process rewards in VPRM\citep{pronesti-etal-2026-beyond}, and tool-supported checking in AgentV-RL\citep{zhang-etal-2026-agentv}. 
We test whether these mechanisms transfer from formula-centered or document-grounded settings to the broader five-stage CFA and FRM distribution.

\paragraph{Financial Agents and Adaptive Reasoning.} Recent systems extend static financial QA toward adaptive tool use and long-horizon workflows, including AlphaQuanter\citep{deng-etal-2026-alphaquanter}, Finch\citep{dong-etal-2026-finch}, FinMaster\citep{jiang-etal-2026-finmaster}, and FinanceHarness\citep{xiao-etal-2026-financeharness}. RouteRAG\citep{guo-etal-2026-routerag} and Reflective RAG\citep{wu-etal-2026-reflective} further study adaptive retrieval, revision, and stopping. 
These directions complement our controlled setting, but open-ended tool use and variable trajectories complicate item-level comparison and may expose sequestered questions. We therefore use fixed-budget, closed-world interventions and leave adaptive agentic control as a future direction motivated by the remaining retrieval and verification failures.

\textsc{FinExam-10K} therefore connects curriculum-scale benchmarking with controlled intervention analysis, asking not only which models pass but also which errors external knowledge can repair and which remain unresolved.

\section{Dataset}
\label{sec:dataset}

\textsc{FinExam-10K} comprises a Full-Coverage Track of 10{,}198 English multiple-choice questions aligned with CFA Levels I--III and FRM Parts I--II. It is the largest reported English benchmark dedicated to professional financial examination reasoning.\footnote{The literature search was done on July 31, 2026 and dedicated to English-language financial examination reasoning.} Every item contains a stem, options, a gold answer, a reference rationale, and the examination program and stage. CFA questions have three options, whereas FRM questions have four.

Every record is structurally complete. Context completeness records a different property: whether a standalone item preserves all answer-necessary material from its source set. Some questions originate in multi-item vignettes or exhibit-based cases. When such cases are flattened, the shared text, table, image, or figure may no longer be attached to an individual subquestion. A context-incomplete flag therefore does not indicate malformed JSON or a missing core field.

\subsection{Construction and Expert Reannotation}
\label{sec:expert-review}
\label{sec:quality}

We construct the benchmark through rationale filtering, normalization, global deduplication, and two-stage expert reannotation. A four-member finance-qualified team reviews all 10{,}198 retained records. In the first stage, three external reviewers receive subsets matched to their financial training and CFA or FRM experience. They check and, where necessary, revise the stem, answer options, gold answer, source-provided rationale, examination stage, available subject label, and associated metadata. In the second stage, the lead curator reviews every retained item, integrates corrections, and applies deterministic checks for schema validity, identifier uniqueness, answer-option mapping, and content duplication.

The fixed 17-model response matrix and empirical difficulty labels are frozen only after both review stages are complete. Where a referenced exhibit or shared vignette could be recovered faithfully from the source, it was restored. We did not generate or infer missing evidence. 
Residual context-incomplete cases are flagged after the difficulty freeze. This flag does not alter the question, gold answer, access partition, or empirical band. Appendix~\ref{app:expert-review} documents the curation pipeline, reviewer assignments, compensation, and reannotation protocol.

\begin{table}[t]
\centering
\small
\setlength{\tabcolsep}{4.2pt}
\renewcommand{\arraystretch}{1.06}
\begin{tabular}{@{}llrrr@{}}
\toprule
\textbf{Dimension} & \textbf{Category}
& \textbf{All} & \textbf{Public} & \textbf{Held-out} \\
\midrule
\multirow{7}{*}{Stage}
& CFA Level I   & 4{,}763 & 3{,}109 & 1{,}654 \\
& CFA Level II  & 2{,}215 & 1{,}030 & 1{,}185 \\
& CFA Level III & 653     & 179     & 474 \\
& \textbf{CFA total} & \textbf{7{,}631} & \textbf{4{,}318} & \textbf{3{,}313} \\
\cmidrule(lr){2-5}
& FRM Part I    & 1{,}652 & 516     & 1{,}136 \\
& FRM Part II   & 915     & 276     & 639 \\
& \textbf{FRM total} & \textbf{2{,}567} & \textbf{792} & \textbf{1{,}775} \\
\midrule
Access
& \textbf{All questions} & \textbf{10{,}198} & \textbf{5{,}110} & \textbf{5{,}088} \\
\midrule
\multirow{3}{*}{\shortstack[l]{Difficulty\\Level}}
& Easy   & 6{,}578 & 3{,}164 & 3{,}414 \\
& Medium & 2{,}183 & 1{,}087 & 1{,}096 \\
& Hard   & 1{,}437 & 859     & 578 \\
\bottomrule
\end{tabular}
\caption{Composition of \textsc{FinExam-10K}. Public items are
identified by \texttt{Mock} or \texttt{Practice Exam} source
metadata. Difficulty is derived from Section~\ref{sec:difficulty}.}
\label{tab:composition}
\end{table}

\subsection{Public and Held-Out Evaluation}
\label{sec:dataset-partitions}

The questions derive from CFA-aligned and FRM-aligned preparatory materials available for academic research. Official CFA Institute and GARP examination content is excluded. The 5{,}110 items labeled \texttt{Mock} or \texttt{Practice Exam} form the public partition and are released in structured JSON format. 
The remaining 5{,}088 items form a held-out partition evaluated through the leaderboard protocol. Both partitions follow the same filtering, normalization, deduplication, expert reannotation, answer extraction, and scoring. The held-out partition limits repeated test exposure and supports controlled model comparison. Its access status is not used to infer that an item is an official examination question. Appendix~\ref{app:partition-audit} reports the partition audit and ranking-stability analysis.

\paragraph{Held-out curriculum taxonomy.} All 5{,}088 held-out items include curriculum subject metadata. 
Appendix Figure~\ref{fig:subject-taxonomy} presents the complete stage-by-subject map.

\subsection{Empirical Difficulty}
\label{sec:difficulty}

CFA and FRM stages describe curriculum progression for human candidates, but they do not define a common item-level scale across the two programs. We therefore derive empirical difficulty from one frozen prediction per model on all 10{,}198 items. Let $c_{m,i}\in\{0,1\}$ indicate whether model $m$ answers item $i$ correctly. The panel contains ten API-served models and seven locally evaluated models. To prevent the larger group from receiving mechanically greater influence, we average correctness within each access group and weight the two means equally:

\begin{equation}
s_i = \frac{1}{2}
\left(
\frac{1}{10}\sum_{m\in\mathcal{M}_{\mathrm{API}}} c_{m,i}
+
\frac{1}{7}\sum_{m\in\mathcal{M}_{\mathrm{local}}} c_{m,i}
\right).
\label{eq:main-dataset-consensus}
\end{equation}

This grouping is a balancing device rather than a taxonomy of model capability. The frozen empirical band is

\begin{equation}
d_i=
\begin{cases}
\textsc{Easy},   & s_i \geq 2/3,\\
\textsc{Hard},   & s_i \leq 1/3,\\
\textsc{Medium}, & \text{otherwise}.
\end{cases}
\label{eq:main-dataset-band}
\end{equation}

The resulting version contains 6{,}578 Easy, 2{,}183 Medium, and 1{,}437 Hard items. Difficulty-conditioned results use leave-one-model-out bands, so the evaluated model does not help define its own test subset. Appendix~\ref{app:difficulty} reports alternative weighting rules, threshold sensitivity, split-half reliability, and leave-one-model-out stability.

\paragraph{Complementary evaluation tracks.}
The 10{,}198-item \emph{Full-Coverage Track} preserves the complete curated examination-aligned item universe and is used for complete coverage, difficulty construction, and the maintained leaderboard. The 7{,}625-item \emph{Context-Complete Reasoning Track} contains items whose answer-necessary local record is complete and is the primary evidence base for financial reasoning from that supplied record.  
The reasoning track is a fixed filter rather than a second difficulty construction: public and held-out membership remain unchanged, every retained item inherits its frozen Easy, Medium, or Hard label, and the labels are not recomputed. It contains 372 Hard items. Correctness on the remaining 2{,}573 items cannot be interpreted solely as grounded reasoning from the supplied local record. Appendix~\ref{app:context-completeness-audit} reports the audit, labeler calibration, subset composition, and complete sensitivity results.

\subsection{Validation and Versioning}
\label{sec:dataset-validation}

The empirical ordering is stable across alternative panel weightings. The adopted rule agrees with the flat 17 LLMs mean on 95.8\% of items, with Spearman \(\rho=0.998\), and with an equal-weight three family rule on 90.8\% of items, with \(\rho=0.991\). The Spearman--Brown corrected split-half reliability is \(0.902\), and the minimum leave-one-model-out correlation is \(0.983\). The association between empirical difficulty and examination stage is modest at Cram\'er's 
\(V=0.174\), showing that the bands do not merely reproduce curriculum ordering. Full statistical procedures appear in Appendix~\ref{app:difficulty}.

\textsc{FinExam-10K} follows a versioned quarterly refresh protocol
with maintenance checkpoints in February, May, August, and November.

\section{Evaluation Framework}
\label{sec:evaluation-framework}

We evaluate \textsc{FinExam-10K} in two complementary tracks and a matched intervention study. The Full-Coverage Track supports direct comparison of 17 frozen models, difficulty construction, and leaderboard reporting. The Context-Complete Reasoning Track is the primary evidence base for reasoning from the supplied local record. Across intervention conditions, the question, answer options, backbone, decoding configuration, answer extraction, and scoring rule remain fixed.

\subsection{Models and Common Protocol}
\label{sec:evaluated-models}

The panel contains 10 \textit{Closed-source Proprietary Models}, 2 \textit{Open-weight Reasoning Models}, and 5 \textit{Finance-specialized Models}.
Every model answers the same 10{,}198 Full-Coverage Track questions without gold labels or reference rationales. 
The frozen predictions define the leaderboard and empirical-difficulty matrix, with leave-one-model-out bands used for model-specific reporting.

For intervention experiments, we use DeepSeek-R1 with CoT and GPT-4o with PoT, following the strongest corresponding settings in FinanceReasoning~\citep{tang-etal-2025-financereasoning}. We use deterministic decoding where supported, apply a fixed answer-extraction cascade, and score unparseable or length-truncated outputs as incorrect, preserving a denominator of 10{,}198. Appendix~\ref{app:reproducibility} provides model revisions, prompts, hardware, and software details.

\subsection{Controlled Knowledge Interventions}
\label{sec:controlled-interventions}

\paragraph{Direct and Function-RAG.}
Direct supplies no external function knowledge. Function-RAG transfers the chain-specific FinanceReasoning baselines~\citep{tang-etal-2025-financereasoning}. 
For GPT-4o PoT, a generated query retrieves the Contriever top 30 and an LLM relevance judge retains zero to three functions. 
For DeepSeek-R1 CoT, a BM25 top-30 candidate set is followed by the transferred LLM-instructed retrieval and judging stage. 
%A raw BM25 CoT arm is reported only as an appendix ablation.

\paragraph{FunctionGraph-RAG and verification.}
The two reasoning chains use separate frozen selectors because their retrieval and graph constructions differ. For GPT-4o PoT, the Contriever top-30 candidates are expanded by one hop over a static graph containing two edge types: shared \texttt{article\_title} and Contriever $k$-nearest-neighbor links with $k=4$. A four-feature linear pairwise ranker, trained on 511 reachable FinanceReasoning Easy and Medium labels, returns at most three functions. For DeepSeek-R1 CoT, BM25 top-30 candidates are expanded for up to two hops over a graph linking functions that share a normalized input or output quantity, with the pool capped at 80. A 56-feature linear reranker, trained on 890 reachable Easy, Medium, and Hard labels, returns the top ten. 
Neither selector uses a FinExam-10K item, answer, rationale, correctness signal, or model output, and both are transferred without adaptation. Appendix~\ref{app:functiongraph-rag-details} records the exact features, training partitions, and feature patterns.

For GPT-4o, \textsc{FunctionGraph-RAG+Verifier} is applied when Function-RAG and FunctionGraph-RAG disagree. The informed verifier receives the item, options, both generated programs or traces, and both execution results, then selects one frozen branch answer. It is a bounded post-generation check rather than an iterative repair process. %The later post-execution self-check is a separate diagnostic.

\subsection{Direct-Conditioned Invocation Gate}
\label{sec:selective-router}

Static augmentation can repair a Direct error while disrupting an answer that was already correct. We therefore train a one-step gate that runs Direct first and then decides whether FunctionGraph-RAG should be invoked. From the item and completed Direct call, it constructs a 27-dimensional feature vector $\mathbf{x}_i$ covering item form, Direct parsing and execution status, token and latency summaries, error indicators, and the Direct option. It excludes gold labels, reference rationales, retrieval state, FunctionGraph-RAG outputs, and all cross-branch features.

\begin{equation}
\begin{aligned}
r_i &= \mathbb{I}\!\left[p_\theta(\mathbf{x}_i) \geq 0.68\right],\\
\hat{y}^{\,\mathrm{gate}}_i &=
\begin{cases}
\hat{y}^{\,\mathrm{dir}}_i, & r_i=0,\\
\hat{y}^{\,\mathrm{fg}}_i, & r_i=1.
\end{cases}
\end{aligned}
\label{eq:router-decision}
\end{equation}

Training uses only the 5{,}110 public items. The positive target marks cases where FunctionGraph-RAG uniquely corrects a Direct error. Correctness disagreements receive weight 1.0 and ties weight 0.15. Five-fold stratified cross-validation with single seed selects the logistic-regression regularization $C=0.5$ and threshold 0.68 before held-out evaluation.

For evaluation, the frozen policy is applied offline to precomputed Direct and FunctionGraph-RAG predictions, but the decision function consumes only item and Direct features. The same policy can therefore be deployed lazily, with FunctionGraph-RAG executed only after a trigger. We report $1+$ trigger rate as the \emph{implied branch-call count}; it is not measured latency, token usage, monetary cost, or energy. Three selectors that inspect several completed branches are retained as higher-cost appendix ablations.

\subsection{Metrics and Statistical Testing}
\label{sec:intervention-metrics}

We report accuracy and matched transitions from Direct. A \emph{rescue} changes an incorrect Direct answer to correct, whereas a \emph{harm} changes a correct Direct answer to incorrect. With counts $R$ and $H$,
\begin{equation}
\Delta = 100\frac{R-H}{N}.
\label{eq:net-repair}
\end{equation}
We use two-sided exact McNemar tests on $(R,H)$. 
A separate post hoc oracle ceiling uses gold correctness after all branch outputs are frozen; it is nondeployable and never used by the gate. 
Static interventions are reported on both tracks; supplied-record conclusions use the Context-Complete Reasoning Track. The gate is trained on public data and evaluated on the 5{,}088 held-out items and their 4{,}219-item context-complete intersection.

\section{Results and Analysis}
\label{sec:results}

\paragraph{RQ1: Where do current LLMs succeed and fail across the CFA and FRM curriculum?}

\textbf{Overall and stage-level performance.} Table~\ref{tab:rq1-stage-difficulty-results} shows a clear advantage for frontier proprietary models. Gemini-3.1-Pro and GPT-5.6-Sol lead overall at 85.29\% and 84.75\%, compared with 75.76\% for the strongest open-weight reasoning model and 68.07\% for the strongest finance-specialized model. Yet every model scores lower on CFA Level II than on Level III, showing that curriculum stage alone does not determine item difficulty.

\begin{table*}[t]
\centering
\footnotesize
\setlength{\tabcolsep}{2.4pt}
\renewcommand{\arraystretch}{1.08}
\begin{tabularx}{0.92\textwidth}{@{}l*{9}{>{\raggedleft\arraybackslash}X}@{}}
\toprule
\multirow{2}{*}{\textbf{Model}}
& \multicolumn{3}{c}{\textbf{CFA}}
& \multicolumn{2}{c}{\textbf{FRM}}
& \multicolumn{3}{c}{\textbf{Difficulty band}}
& \multirow{2}{*}{\textbf{All}} \\
\cmidrule(lr){2-4}
\cmidrule(lr){5-6}
\cmidrule(lr){7-9}
& \textbf{L-I} & \textbf{L-II} & \textbf{L-III}
& \textbf{P-I} & \textbf{P-II}
& \textbf{Easy} & \textbf{Med.} & \textbf{Hard} & \\
\midrule
\multicolumn{10}{@{}l}{\textit{Proprietary models}} \\
Gemini-3.1-Pro
& \textbf{90.95} & \textbf{71.74} & \underline{77.79} & \underline{90.86} & \underline{83.93}
& \cellcolor{green!98!red!18}\underline{97.92} & \cellcolor{green!82!red!18}\textbf{82.13} & \cellcolor{green!35!red!18}\textbf{34.68} & \textbf{85.29} \\
GPT-5.6-Sol
& \underline{89.61} & \underline{71.15} & \textbf{78.10} & \textbf{91.28} & \textbf{85.36}
& \cellcolor{green!98!red!18}\textbf{98.09} & \cellcolor{green!82!red!18}\underline{81.97} & \cellcolor{green!30!red!18}\underline{30.14} & \underline{84.75} \\
GPT-5.6-Terra
& 86.73 & 69.21 & 75.80 & 90.38 & 83.72
& \cellcolor{green!97!red!18}97.23 & \cellcolor{green!76!red!18}76.23 & \cellcolor{green!26!red!18}26.36 & 82.55 \\
DeepSeek-V4-Pro
& 87.05 & 69.03 & 75.65 & 89.47 & 82.62
& \cellcolor{green!97!red!18}97.32 & \cellcolor{green!75!red!18}75.05 & \cellcolor{green!27!red!18}26.63 & 82.40 \\
GPT-5.6-Luna
& 85.53 & 65.96 & 72.89 & 86.74 & 79.13
& \cellcolor{green!97!red!18}96.73 & \cellcolor{green!70!red!18}69.72 & \cellcolor{green!20!red!18}19.82 & 80.09 \\
Claude-Sonnet-5
& 84.95 & 66.05 & 73.81 & 86.74 & 78.25
& \cellcolor{green!95!red!18}95.05 & \cellcolor{green!72!red!18}71.54 & \cellcolor{green!23!red!18}23.02 & 79.82 \\
DeepSeek-R1
& 84.90 & 65.55 & 74.89 & 86.02 & 79.45
& \cellcolor{green!97!red!18}96.87 & \cellcolor{green!69!red!18}68.60 & \cellcolor{green!18!red!18}18.28 & 79.75 \\
Qwen3.7-Max
& 80.54 & 61.58 & 70.75 & 78.21 & 73.01
& \cellcolor{green!92!red!18}91.91 & \cellcolor{green!58!red!18}57.86 & \cellcolor{green!20!red!18}20.15 & 74.74 \\
GPT-5.5
& 79.00 & 58.10 & 66.31 & 81.90 & 75.96
& \cellcolor{green!91!red!18}90.85 & \cellcolor{green!54!red!18}53.98 & \cellcolor{green!24!red!18}23.54 & 73.85 \\
GPT-4o
& 75.96 & 56.75 & 63.09 & 70.52 & 65.57
& \cellcolor{green!88!red!18}87.68 & \cellcolor{green!46!red!18}45.75 & \cellcolor{green!15!red!18}15.42 & 69.15 \\
\addlinespace[2pt]
\multicolumn{10}{@{}l}{\textit{Open-weight reasoning models}} \\
GPT-OSS-120B
& 81.21 & 61.81 & 71.36 & 83.96 & 69.51
& \cellcolor{green!95!red!18}94.85 & \cellcolor{green!59!red!18}59.36 & \cellcolor{green!16!red!18}16.09 & 75.76 \\
GPT-OSS-20B
& 75.50 & 57.52 & 64.93 & 77.18 & 62.95
& \cellcolor{green!89!red!18}88.56 & \cellcolor{green!50!red!18}49.52 & \cellcolor{green!16!red!18}16.15 & 70.06 \\
\addlinespace[2pt]
\multicolumn{10}{@{}l}{\textit{Finance-specialized models}} \\
ODA-Fin-RL-8B
& 75.46 & 54.49 & 62.63 & 72.22 & 58.91
& \cellcolor{green!87!red!18}87.13 & \cellcolor{green!45!red!18}44.78 & \cellcolor{green!14!red!18}14.09 & 68.07 \\
Fin-o1-14B
& 72.43 & 53.23 & 59.11 & 69.07 & 60.44
& \cellcolor{green!84!red!18}83.65 & \cellcolor{green!40!red!18}40.23 & \cellcolor{green!19!red!18}18.96 & 65.79 \\
DianJin-R1-32B
& 70.65 & 52.42 & 60.49 & 68.46 & 62.30
& \cellcolor{green!84!red!18}83.51 & \cellcolor{green!40!red!18}39.81 & \cellcolor{green!13!red!18}12.63 & 64.93 \\
Hawkish-8B
& 57.65 & 45.33 & 49.62 & 51.51 & 44.04
& \cellcolor{green!64!red!18}64.23 & \cellcolor{green!36!red!18}36.03 & \cellcolor{green!18!red!18}17.86 & 52.25 \\
Fin-R1-7B
& 57.74 & 40.90 & 47.47 & 52.91 & 45.36
& \cellcolor{green!66!red!18}66.38 & \cellcolor{green!25!red!18}25.48 & \cellcolor{green!14!red!18}14.15 & 51.53 \\
\bottomrule
\end{tabularx}
\caption{
Accuracy (\%) on the 10{,}198-item Full-Coverage Track by examination stage and leave-one-model-out difficulty band. Best and second-best results in each column are shown in bold and underlined, respectively.
}
\label{tab:rq1-stage-difficulty-results}
\vspace{-0.45em}
\end{table*}

\begin{figure}[!t]
    \centering
    \includegraphics[width=\columnwidth]{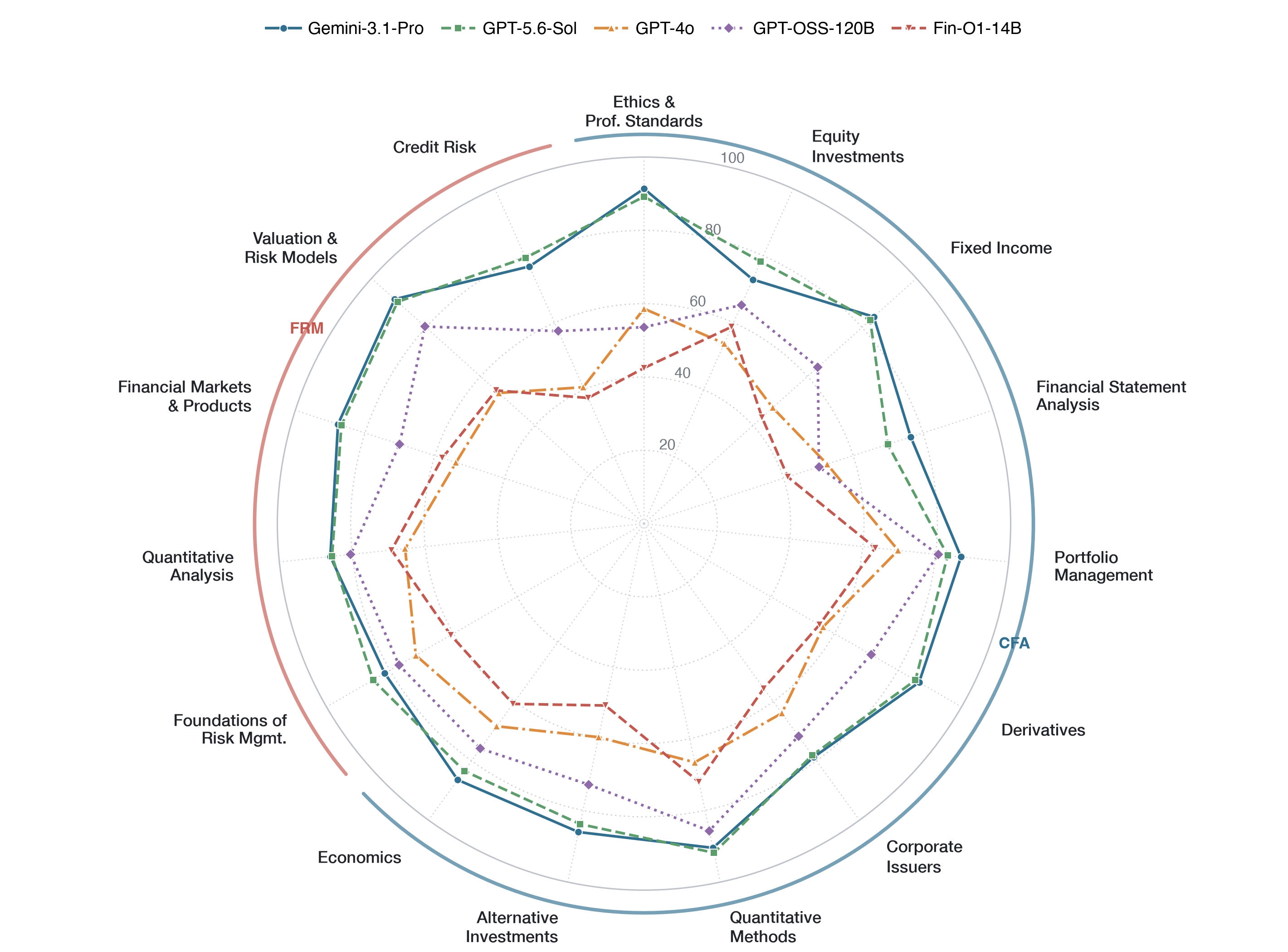}
    \caption{Chance normalized accuracy across 15 CFA and FRM subjects in the held-out partition. Only subjects with at least 150 labeled items are included.
    }
    \label{fig:subject-radar}
    \vspace{-0.45em}
\end{figure}

\textbf{Difficulty and subject heterogeneity.} On the Full-Coverage Track, leading models exceed 97\% on Easy items but reach only 34.68\% on Hard items. Since 1{,}065 of the 1{,}437 Hard records lack locally attached parent evidence, this is a mixed coverage-and-reasoning challenge. On the 372-item context-complete Hard subset, the best score rises to 54.57\%, yet only three models exceed its 30.40\% item-weighted chance baseline. Context loss therefore amplifies, but does not explain, the Hard-band collapse. CFA Level III and FRM Part II are enriched by factors of 1.93 and 1.98. Figure~\ref{fig:subject-radar} shows further subject heterogeneity, while the full Context-Complete Reasoning Track preserves the model ordering ($\rho=0.988$; Appendix~\ref{app:context-complete-track}).

\textbf{Shared failures.} Among the 369 context-complete Hard items for which the concentration statistic is defined, the median share of erroneous votes assigned to one distractor is 0.923, compared with an exact item-specific null median of 0.605. The observed concentration exceeds its null on 346 items, and all erroneous votes select one distractor on 148 (\(p=1.9\times10^{-75}\), one-sided exact sign test). A stricter universal-failure core contains 188 items missed by all 17 LLM models. Among them, 141 are context incomplete and 47 are context complete; on 41 items, every model selects the same wrong option.
Answer position provides an additional source of systematic variation. The universal-failure core overrepresents option A among gold labels (44.7\% versus 29.9\% benchmark-wide), while model predictions favor B (39.0\%). The context-complete Hard subset exhibits a different pattern, with under-selection of A and over-selection of C. Position preference may therefore contribute to measured failure, but its direction is slice-specific rather than a single benchmark-wide bias. Appendix~\ref{app:rq1-diagnostics} provides the slice definitions, null construction, model results, and position distributions.

\paragraph{RQ2: Under what conditions do static knowledge interventions help or harm financial reasoning?}

\textbf{Static retrieval creates substantial movement, but no reliable aggregate gain.} Table~\ref{tab:rq2-full} shows that the FinanceReasoning Function-RAG setting does not transfer uniformly~\citep{tang-etal-2025-financereasoning}. Under DeepSeek-R1 CoT, Function-RAG yields 505 rescues and 538 harms (\(-0.32\) points, \(p=.322\)); FunctionGraph-RAG yields 509 and 500 (\(+0.09\), \(p=.801\)). Under GPT-4o PoT, the two retrieval variants decline by 1.27 and 1.30 points. The verifier removes 130 harms at the cost of 43 rescues, recovering 0.86 points but remaining 0.45 below Direct (\(p=.158\)). The pattern persists on the Context-Complete Reasoning Track. Oracle gains of 7.25 and 7.46 points confirm branch complementarity, so the bottleneck is preserving correct Direct answers rather than finding repairable errors (Appendix~\ref{app:rq2-diagnostics}).

\textbf{The relevance judge separates opposing intervention regimes.}
Table~\ref{tab:rq2-function-count} shows that the judge retains no function for 68.7\% of items. Forcing the three-function PoT graph in this stratum reduces accuracy by 1.10 points (\(p=.0040\)). When the judge retains exactly one relevant function, graph expansion instead improves accuracy by 2.92 points (\(p=.0007\)).
Both effects remain significant after Benjamini--Hochberg correction, whereas the two larger-count strata show no reliable difference. Because the PoT graph branch supplies three functions throughout, these comparisons jointly vary judged evidence need and injection volume rather than isolating graph topology. The count-matched three-function stratum likewise shows no significant gain. The judge's abstention is therefore informative. Forced injection is harmful when no relevant function is identified, while the clearest benefit appears when graph expansion begins from one plausible seed.

\begin{table}[!t]
\centering
\scriptsize
\setlength{\tabcolsep}{2.4pt}
\renewcommand{\arraystretch}{1.06}
\begin{tabularx}{\columnwidth}{
@{}>{\raggedright\arraybackslash}X r r r r r@{}
}
\toprule
\textbf{Method}
& \textbf{Acc.}
& \textbf{Resc.}
& \textbf{Harm}
& \textbf{$\Delta$}
& \textbf{$p$} \\
\midrule
\multicolumn{6}{@{}l}{\textit{CoT, DeepSeek-R1}} \\
Direct & 79.75 & -- & -- & -- & -- \\
Function-RAG & 79.43 & 505 & 538 & $-0.32$ & .322 \\
FunctionGraph-RAG & \textbf{79.84} & 509 & 500 & $+0.09$ & .801 \\
\midrule
\multicolumn{6}{@{}l}{\textit{PoT, GPT-4o}} \\
Direct & \textbf{69.37} & -- & -- & -- & -- \\
Function-RAG & 68.09 & 427 & 557 & $-1.27$ & $<.001$ \\
FunctionGraph-RAG & 68.06 & 527 & 660 & $-1.30$ & $<.001$ \\
\quad + Branch Verifier & 68.92 & 484 & 530 & $-0.45$ & .158 \\
\bottomrule
\end{tabularx}
s\caption{Intervention outcomes on the Full-Coverage Track ($N=10{,}198$). Rescue and harm are transitions relative to Direct.}
\label{tab:rq2-full}
\vspace{-0.45em}
\end{table}

\begin{table}[!t]
\centering
\scriptsize
\setlength{\tabcolsep}{3.0pt}
\renewcommand{\arraystretch}{1.06}
\begin{tabular*}{\columnwidth}{
@{\extracolsep{\fill}}c r r r r r@{}
}
\toprule
\textbf{\#Fn} & \textbf{$N$} & \textbf{Fn-RAG} & \textbf{FG-RAG} & \textbf{$\Delta$} & \textbf{$p$} \\
\midrule
0 & 7{,}002 & \textbf{72.08} & 70.98 & $-1.10$ & .0040 \\
1 & 2{,}225 & 60.31 & \textbf{63.24} & $+2.92$ & .0007 \\
2 & 479 & 55.53 & \textbf{56.99} & $+1.46$ & .483 \\
3 & 492 & 58.74 & \textbf{59.15} & $+0.41$ & .897 \\
\bottomrule
\end{tabular*}
\caption{GPT-4o PoT conditioned on the number of functions retained by the Function-RAG judge. }
\label{tab:rq2-function-count}
\vspace{-0.45em}
\end{table}

\paragraph{RQ3: Can a low-cost gate learn when FunctionGraph-RAG should be invoked?} RQ2 shows that structured augmentation benefits only a subset of questions and can be harmful when applied indiscriminately.
We therefore select a Direct-conditioned gate using only item- and Direct-stage features. Its configuration and threshold are chosen solely by five-fold out-of-fold performance on the public partition and frozen before held-out evaluation.
On the 5{,}088 held-out items, the gate triggers on 404 questions (7.9\%) and raises accuracy from 70.83\% to 71.23\% (\(55\) rescues, \(35\) harms, \(+0.39\) points, \(p=.0446\)). Under lazy execution, this corresponds to an implied \(1.08\times\) branch-call count. On the 4{,}219-item held-out context-complete subset, it triggers on 4.3\% of questions and gains \(0.55\) points (\(p=.0003\)). Thus, most questions should remain on the Direct path, while selective invocation recovers part of the graph branch's conditional value.
Exploratory bounded probes yield no further paired gain from same-backbone verification (\(+0.27\) points), or plan-then-solve (\(-3.33\) points); answerability detection carries useful risk signal but rejects 53.3\% of answerable items. Full details in Appendix~\ref{app:rq3-gate}. A future financial reasoning agent would move this decision upstream and make it iterative, choosing among retrieval, graph expansion, execution, verification, revision, and stopping.

\section{Conclusions}

We introduced \textsc{FinExam-10K}, an expert-reviewed benchmark of 10{,}198 questions across all five CFA and FRM stages, with 5{,}110 public items, 5{,}088 sequestered items, a quarterly maintained leaderboard, and a fixed 17-model panel. 
The 7{,}625-item Context-Complete Reasoning Track confirms that the principal rankings and intervention conclusions persist when all answer-necessary evidence is locally attached. 
Static Function-RAG and chain-specific FunctionGraph-RAG repair many errors but introduce comparable harm. A public-trained Direct-conditioned gate recovers a modest held-out gain while invoking the graph branch on only 7.9\% of items. 
Together, the benchmark and matched diagnostics show that progress requires selective access to independent evidence and tools, rather than uniformly adding retrieval or another pass over the same reasoning.

\section*{Limitations and Future Work}

\paragraph{Detached parent context.}
Every record contains a stem, options, a gold answer, and a rationale, but 2{,}573 standalone records do not retain evidence supplied at the parent item-set level, such as a shared vignette, table, image, or exhibit. The Full-Coverage Track therefore measures the released itemized representation and may combine financial reasoning with prior exposure, reconstruction from partial cues, or chance. Claims about reasoning from supplied evidence rely primarily on the Context-Complete Reasoning Track. We retain both tracks to preserve curriculum coverage without conflating representation effects with grounded reasoning, and future releases will preserve vignette identifiers and shared-evidence bundles more systematically.

\paragraph{Difficulty and answer position.}
The empirical bands are defined by a frozen model panel rather than human psychometric calibration. We also observe slice-specific answer-position effects. Future versions will add option-permutation tests and, where available, human response evidence.

\paragraph{Intervention and agentic scope.}
The PoT and CoT FunctionGraph-RAG results use different frozen selectors and support only within-chain matched comparisons. The PoT judge-count analysis also varies evidence volume outside the count-matched stratum. Gate efficiency is an implied branch-call count, not measured latency, tokens, energy, or monetary cost. Finally, the verification, abstention, and planning probes are single-backbone, fixed-budget diagnostics rather than a learned multi-step agent. A fuller agent would require independent evidence, grounded tool arguments, execution-aware revision, and calibrated stopping.

\section*{Ethical Statement}

\paragraph{Data provenance and release.}
\textsc{FinExam-10K} is constructed from CFA-aligned and FRM-aligned preparatory materials available for academic research. Official CFA Institute and GARP examination content is excluded. The public release contains only the 5{,}110 mock and practice items whose source terms permit academic redistribution; the remaining 5{,}088 items are kept sequestered and accessed only through the evaluation protocol. CFA and FRM are used descriptively to identify curriculum alignment. This work is not affiliated with or endorsed by CFA Institute or GARP. The benchmark contains no personal financial records, account credentials, or other personally identifiable information.

\paragraph{Expert reannotation.}
All retained items underwent two-stage review by a four-member finance-qualified team. The three external reviewers were recruited for stage-matched expertise, informed that their corrections would be used to construct a research benchmark, and consented to the reporting of their anonymized qualifications and responsibilities. Each external reviewer was compensated at a rate meeting or exceeding the statutory minimum wage applicable in their country of residence. Reviewer identities remain anonymized.

\paragraph{Intended use and broader impact.}
The benchmark is intended for research on financial reasoning, evaluation, retrieval, verification, and tool use. It is not financial advice, a substitute for professional certification, or a resource for reconstructing official examinations. Benchmark performance should not be interpreted as evidence that a model is safe for autonomous investment, risk-management, compliance, or advisory decisions. We report context and answer-position diagnostics and maintain a sequestered partition to make measurement limitations visible and reduce repeated test exposure.

\clearpage
\bibliographystyle{acl_natbib}
\bibliography{main}

\appendix
\clearpage

\section{Dataset Curation and Expert Reannotation}
\label{app:expert-review}

\subsection{Rationale Coverage and Curation Yield}
\label{app:curation-yield}

The initial collection contained 15{,}199 records across the five examination stages. We retained records with a usable reference rationale provided by the source and did not synthesize rationales for records lacking one. This filter retained 13{,}715 records. Normalization and global deduplication then produced 10{,}345 unique rationale-bearing items. Of these, 147 CFA Level III constructed-response items were reserved outside the present multiple-choice benchmark, yielding the final 10{,}198 questions.

\begin{table*}[t]
\centering
\small
\setlength{\tabcolsep}{5.0pt}
\renewcommand{\arraystretch}{1.08}
\begin{tabular}{@{}lrrrrrr@{}}
\toprule
\textbf{Examination Stage}
& \textbf{Raw}
& \textbf{With Rationale}
& \textbf{Rationale Coverage}
& \textbf{Post-Dedup.}
& \textbf{Final MCQ}
& \textbf{Final Yield} \\
\midrule
CFA Level I
& 7{,}496 & 7{,}470 & 99.7\% & 4{,}763 & 4{,}763 & 63.5\% \\
CFA Level II
& 2{,}556 & 2{,}529 & 98.9\% & 2{,}215 & 2{,}215 & 86.7\% \\
CFA Level III
& 950 & 950 & 100.0\% & 800 & 653 & 68.7\% \\
\cmidrule(lr){1-7}
\textit{CFA total}
& 11{,}002 & 10{,}949 & 99.5\% & 7{,}778 & 7{,}631 & 69.4\% \\
\midrule
FRM Part I
& 2{,}667 & 1{,}823 & 68.4\% & 1{,}652 & 1{,}652 & 61.9\% \\
FRM Part II
& 1{,}530 & 943 & 61.6\% & 915 & 915 & 59.8\% \\
\cmidrule(lr){1-7}
\textit{FRM total}
& 4{,}197 & 2{,}766 & 65.9\% & 2{,}567 & 2{,}567 & 61.2\% \\
\midrule
\textbf{All}
& \textbf{15{,}199}
& \textbf{13{,}715}
& \textbf{90.2\%}
& \textbf{10{,}345}
& \textbf{10{,}198}
& \textbf{67.1\%} \\
\bottomrule
\end{tabular}
\caption{Curation yield by examination stage. \textbf{Rationale Coverage} is the proportion of raw records containing a usable reference rationale provided by the source. \textbf{Post-Dedup.} reports rationale-bearing records remaining after normalization and global deduplication. The difference between the CFA Level III post-deduplication and final counts consists of 147 constructed-response items reserved outside the present multiple-choice benchmark. \textbf{Final Yield} is the proportion of raw records retained in the final benchmark.}
\label{tab:curation-yield}
\end{table*}

The CFA collection decreased from 11{,}002 raw records to 7{,}631 final questions, whereas the FRM collection decreased from 4{,}197 to 2{,}567. Their respective retention rates were 69.4\% and 61.2\%. A Pearson chi-square test of retained versus excluded records by examination family gives \(\chi^2(1)=92.47\), \(p=6.83\times10^{-22}\). The final 7{,}631 to 2{,}567 composition therefore partly reflects differential rationale availability and screening retention, rather than only the prevalence of material in the underlying curricula.

\subsection{Reviewer Recruitment, Compensation, and Consent}

The review team comprised one author serving as lead curator and three external reviewers recruited through academic and professional networks. External reviewers were selected for formal training in finance or a closely related quantitative field, direct experience with the assigned CFA or FRM stage, and the ability to inspect structured records using the review interface.

Each external reviewer was compensated at a rate that met or exceeded the statutory minimum wage applicable in their country of residence at the time of annotation. All reviewers were informed that their corrections would be used to construct a research benchmark and consented to the reporting of their anonymized qualifications and responsibilities.

\subsection{Reviewer Profiles}
\label{app:reviewer-profiles}

\paragraph{R1, Lead Curator.}
R1 is a PhD student in computer science with a bachelor's degree in finance and a master's degree in data science. R1 had passed CFA Level I and had two years’ work experience at a leading Asian commercial bank before beginning doctoral research. R1 coordinated collection, normalization, deduplication, and final integration. 
R1 also completed the second reannotation pass over all 10{,}198 items and, after the context-completeness audit and labeler calibration, manually rechecked every frozen Hard item to assign its separate context-completeness status.

\paragraph{R2, CFA Level I Reviewer.}
R2 holds a first-class bachelor's degree in financial mathematics from a leading UK university and has passed CFA Level I. Their training covers financial mathematics, quantitative methods, and structured data analysis. R2 reannotated all CFA Level I records.

\paragraph{R3, FRM Reviewer.}
R3 holds undergraduate training in philosophy, politics, and economics and a master's degree in data science from a top UK university. R3 had passed FRM Part I and was preparing for Part II. Prior to the reannotation, R3 had nearly three years’ experience working at a leading risk management firm in Europe. R3 reannotated all FRM Part I and Part II records.

\paragraph{R4, CFA Levels II and III Reviewer.}
R4 holds bachelor's and master's degrees in finance from a leading university in Asia. R4 had passed CFA Level II while preparing for Level III. Their professional background includes approximately four years’ experience at a top commercial bank, including a two-year secondment to Europe. R4 reannotated all CFA Level II and Level III records.

\begin{table}[t]
\centering
\small
\setlength{\tabcolsep}{4.5pt}
\renewcommand{\arraystretch}{1.08}
\begin{tabular}{@{}llr@{}}
\toprule
\textbf{Reviewer} & \textbf{Assigned Stage} & \textbf{\# Items} \\
\midrule
R2 & CFA Level I & 4{,}763 \\
R4 & CFA Level II & 2{,}215 \\
R4 & CFA Level III & 653 \\
R3 & FRM Part I & 1{,}652 \\
R3 & FRM Part II & 915 \\
\midrule
\multicolumn{2}{l}{Stage-matched first passes} & 10{,}198 \\
\bottomrule
\end{tabular}
\caption{Stage-matched assignments for the first reannotation stage. The lead curator subsequently conducted a full second pass over all 10{,}198 items.}
\label{tab:review-assignments}
\end{table}

\subsection{Item-Level Reannotation Protocol}
\label{app:item-review-protocol}

Reviewers used a Jupyter-based interface that rendered each structured record in a standardized format. Reannotation covered the following fields and checks.

\begin{enumerate}
    \item \textbf{Stem completeness and correctness.}
    Reviewers checked whether the stem contained the conditions, figures, exhibits, and context needed to determine an answer. When missing context was recoverable faithfully from original evidence, reviewers restored it. Otherwise, they retained the record with a context completeness flag and did not supplement it with inferred content.

    \item \textbf{Option integrity.}
    Reviewers checked whether the options were complete, distinguishable, correctly formatted, and free from artifacts that disclosed the gold option. They revised defective options where necessary.

    \item \textbf{Gold-answer validity.}
    Reviewers verified that the gold label mapped to exactly one option and that the selected option was financially and numerically correct. Incorrect keys were corrected.

    \item \textbf{Reference-rationale validity.}
    Reviewers checked whether the rationale provided by the source applied an appropriate financial concept, rule, or formula and supported the gold answer without material contradiction. They corrected the retained rationale where necessary.

    \item \textbf{Stage, subject, and metadata validity.}
    Reviewers verified the examination family, stage, subject label when available, source metadata, item format, and remaining record metadata. Missing subject labels were retained as missing rather than inferred without evidence.
\end{enumerate}

Every record underwent the complete two-stage process before model evaluation. In the first stage, the assigned stage-matched reviewer reannotated the item and recorded an \textsc{Accept}, \textsc{Revise}, or \textsc{Reject} decision together with field-level corrections. In the second stage, the lead curator conducted a full pass over every retained item, integrated corrections, and applied deterministic checks for schema validity, identifier uniqueness, answer-option mapping, and content-hash duplication. Disagreements were resolved with the original reviewer before the model response matrix was frozen.

\begin{table}[t]
\centering
\small
\setlength{\tabcolsep}{5.0pt}
\renewcommand{\arraystretch}{1.08}
\begin{tabular}{@{}lrr@{}}
\toprule
\textbf{Primary Reannotation Outcome}
& \textbf{\# Items}
& \textbf{Share} \\
\midrule
Accepted without substantive change & 9{,}730 & 95.41\% \\
Question or option revision          & 221     & 2.17\% \\
Gold-answer correction               & 36      & 0.35\% \\
Rationale correction                 & 109     & 1.07\% \\
Metadata correction                  & 102     & 1.00\% \\
\midrule
\textbf{All retained items}          & \textbf{10{,}198} & \textbf{100.00\%} \\
\bottomrule
\end{tabular}
\caption{Primary expert-reannotation outcome for each retained item. An item recorded as accepted without substantive change still underwent both complete reannotation stages.}
\label{tab:review-outcomes}
\end{table}

\paragraph{Chronology after reannotation.} After both expert stages were complete for all 10{,}198 items, we froze the 17-model response matrix and \texttt{difficulty}. We then drew and manually audited the stratified 90-item context-completeness sample. The resulting manual labels were next used to calibrate the context-completeness labeler. Only after calibration did the lead curator manually recheck all 1{,}437 items in the frozen Hard band to assign a separate context-completeness status for sensitivity analysis. This final audit and classification pass did not modify \texttt{difficulty}.

\subsection{Held-out Curriculum Subject Taxonomy}
\label{app:subject-taxonomy}

All 5{,}088 items in the held-out evaluation partition include curriculum subject metadata. We preserve subject labels at the examination-stage level rather than collapsing identically named subjects across stages, because the corresponding learning objectives, curricular depth, and item distributions differ. The resulting taxonomy contains 37 stage-specific subject nodes corresponding to 26 distinct subject or pathway categories across CFA Levels I--III and FRM Parts I--II.
The public partition is omitted from this taxonomy because curriculum subject labels are not consistently available for those records. Missing labels are retained as missing rather than assigned through automated inference.

\begin{figure*}[t]
    \centering
    \includegraphics[width=0.98\textwidth]
    {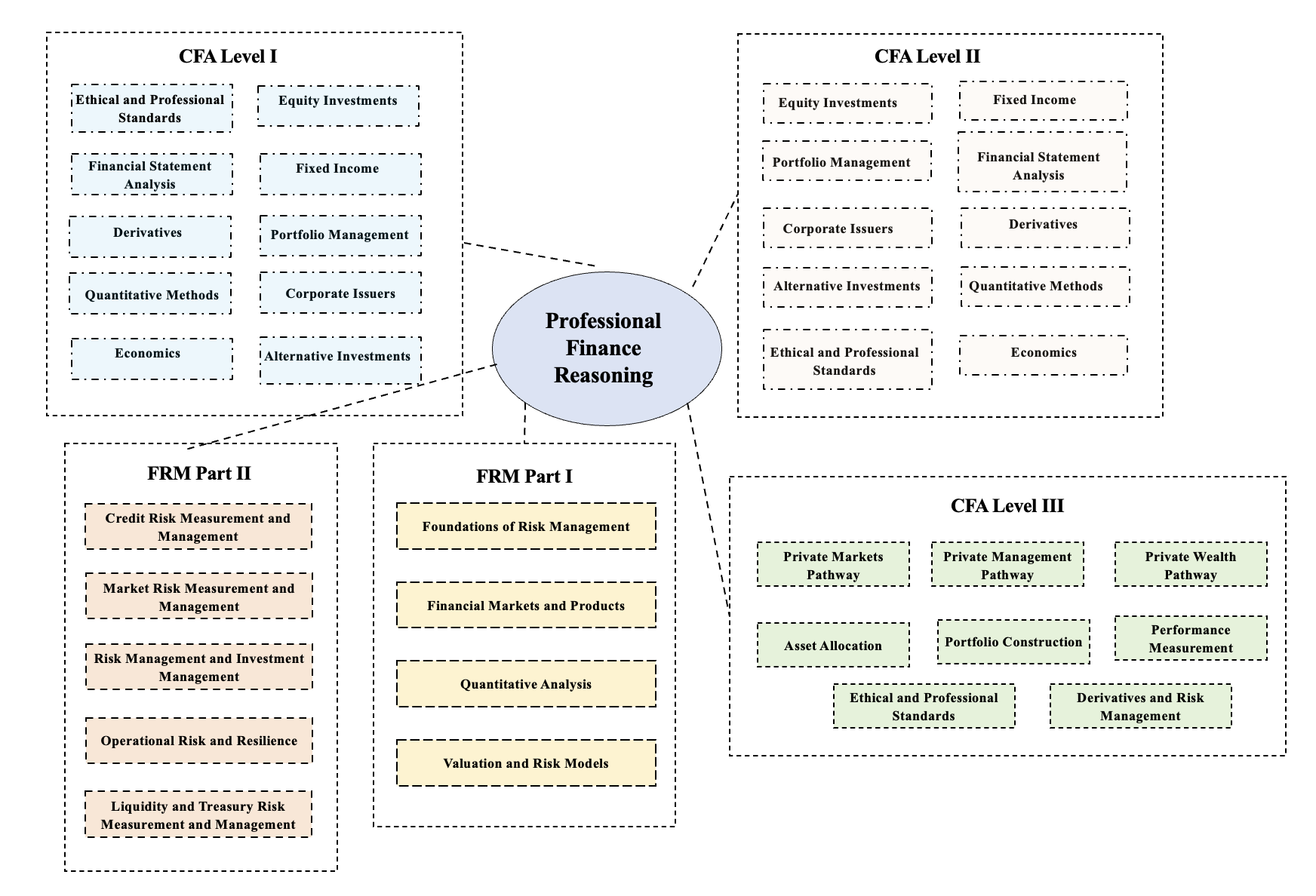}
    \caption{Curriculum subject taxonomy of the 5088-item held-out evaluation partition of FinExam-10K.
    The map contains 37 stage-specific subject nodes spanning
    26 distinct subject or pathway categories across CFA Levels I--III and FRM Parts I--II. Subjects recurring across CFA levels are displayed separately because they correspond to different examination stages, curricular depth, and item distributions.}
    \label{fig:subject-taxonomy}
\end{figure*}

\begin{figure*}[t]
    \centering
    \includegraphics[width=0.96\textwidth,trim=0 18bp 0 0,clip]
    {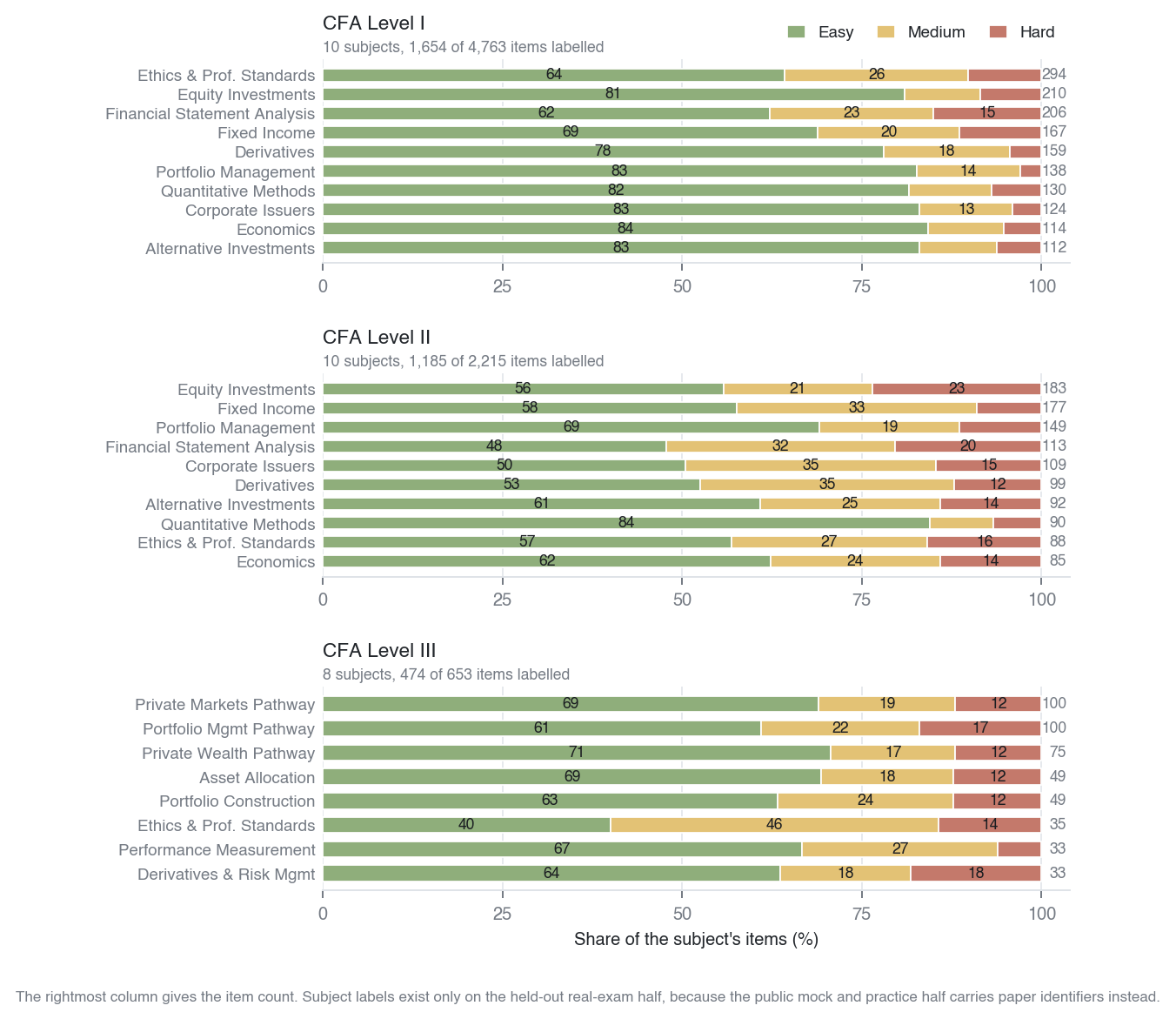}
    \caption{Stage-specific CFA subject composition for the held-out
    evaluation partition of FinExam-10K, the only partition with complete curriculum subject labels. Bars show the share of Easy, Medium, and Hard empirical difficulty bands within each subject; parenthetical and right-margin values give labelled-item counts.}
    \label{fig:subject-composition-cfa}
\end{figure*}

\begin{figure*}[t]
    \centering
    \includegraphics[width=0.96\textwidth,trim=0 18bp 0 0,clip]
    {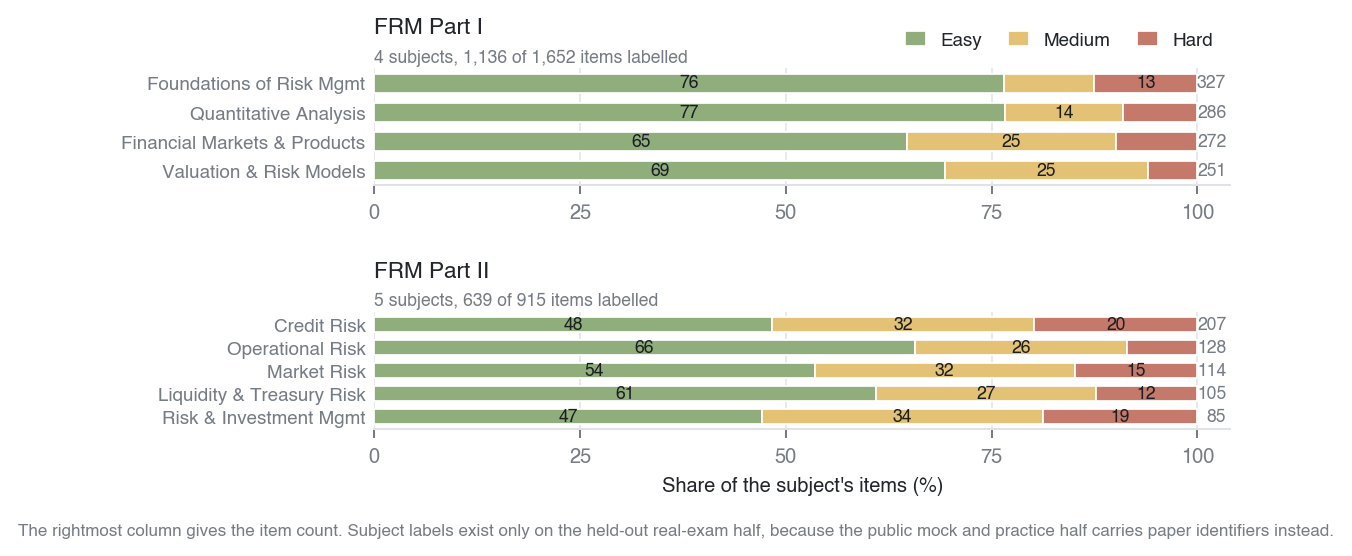}
    \caption{Stage-specific FRM subject composition for the held-out
    evaluation partition of FinExam-10K, the only partition with complete curriculum subject labels. Bars show the share of Easy, Medium, and Hard empirical difficulty bands within each subject; parenthetical and right-margin values give labelled-item counts.}
    \label{fig:subject-composition-frm}
\end{figure*}

Figures~\ref{fig:subject-composition-cfa} and
\ref{fig:subject-composition-frm} show how the frozen empirical difficulty bands are distributed within the held-out curriculum labels.

\section{Partition and Context-Completeness Audits}
\label{app:partition-audit}

\subsection{Operational Partition Audit}

The public partition contains exactly the 5{,}110 records whose source metadata are labeled \texttt{Mock} or \texttt{Practice Exam}. The remaining 5{,}088 records form the held-out partition, are not released, and are evaluated through the leaderboard protocol. Access status is not used to infer provenance. Table~\ref{tab:composition} reports the stage and difficulty composition of both partitions.

\begin{table}[t]
\centering
\scriptsize
\setlength{\tabcolsep}{2.6pt}
\renewcommand{\arraystretch}{1.06}
\begin{tabular}{@{}lrrrr@{}}
\toprule
\textbf{Scope}
& \textbf{Pub.}
& \textbf{Held.}
& \textbf{$\rho$}
& \textbf{P--H} \\
\midrule
Full benchmark & 5{,}110 & 5{,}088 & 0.990 & $-4.58$ \\
Context-complete & 3{,}406 & 4{,}219 & 0.993 & $+2.21$ \\
\bottomrule
\end{tabular}
\caption{Access-partition ranking audit. P--H is the mean
public-minus-held-out accuracy gap in percentage points across the frozen
panel of 17 models.}
\label{tab:access-partition-ranking}
\end{table}

Subject labels are available for all 5{,}088 held-out items. The public items retain their available source metadata, including paper identifiers, but do not have consistently available curriculum subject labels. The reannotation process checked subject labels where they existed and did not impute missing labels.

The Spearman rank correlation between public and held-out model rankings is approximately 0.99 under both full benchmark scoring and scoring on the 7{,}625-item \emph{Context-Complete Reasoning Track}. The public partition can therefore support model development without changing the observed ordering of models in these evaluations.

Ranking consistency must be distinguished from equality in absolute difficulty. Under full-benchmark scoring, all 17 models score lower on the public partition than on the held-out partition. After restriction to the Context-Complete Reasoning Track, all 17 score higher on the public partition than on the held-out partition. For either unanimous direction, the exact two-sided sign test gives
\begin{equation}
p = 2\left(\frac{1}{2}\right)^{17}
  = 1.53\times10^{-5}.
\label{eq:partition-sign-test}
\end{equation}

Context-completeness screening reverses the aggregate public--held-out accuracy gap while leaving the relative model ordering nearly unchanged. The two tracks therefore serve complementary purposes.
The \emph{Full-Coverage Track} preserves the complete curated examination-aligned item universe and supports the maintained leaderboard. The \emph{Context-Complete Reasoning Track} is the primary evidence base for claims about reasoning from the supplied local record. Its results test whether the model rankings, difficulty structure, and intervention conclusions persist after records without locally attached answer-necessary evidence are excluded. We report both tracks rather than treating either as a replacement for the other.

\subsection{Manual Context-Completeness Audit}
\label{app:context-completeness-audit}

Every benchmark record contains a stem, answer options, a gold answer, and a rationale. Context completeness instead evaluates whether the standalone representation contains all answer-necessary material. A record is \emph{context incomplete} when it is detached from answer-necessary shared vignette text or from a referenced table, figure, image, or exhibit. It does not denote malformed JSON or a missing core schema field. Some source cases share one background across several subquestions, so a standalone subquestion can lose that shared context during extraction.

After \texttt{difficulty} was frozen, we manually audited a stratified sample of \(n=90\) items, with 30 items sampled from each frozen difficulty band. This status concerns the current text representation. It does not imply that the original source question was defective or ill-posed.

The audit identified two dominant mechanisms of context loss. In the first category, 636 records lack a referenced table, figure, or exhibit. In the second, 341 records are subquestions detached from a shared vignette or case context when multi-question cases were flattened into separate question-answer records. These counts describe the two identified categories, and they don't overlap.

\begin{table}[t]
\centering
\footnotesize
\setlength{\tabcolsep}{5pt}
\renewcommand{\arraystretch}{1.10}
\begin{tabular}{@{}lcr@{}}
\toprule
\textbf{Band}
& \textbf{Incomplete}
& \textbf{Rate (\%) [Wilson 95\% CI]} \\
\midrule
Easy   & 1/30  & 3.3 [0.6, 16.7]   \\
Medium & 5/30  & 16.7 [7.3, 33.6]  \\
Hard   & 17/30 & 56.7 [39.2, 72.6] \\
\bottomrule
\end{tabular}
\caption{Manual audit of context completeness by frozen empirical difficulty. Thirty items were sampled from each fixed difficulty band. Rates denote the proportion of context-incomplete items, with Wilson 95\% confidence intervals.}
\label{tab:context-completeness-audit}
\end{table}

The band by context-completeness contingency table gives
\begin{equation}
\chi^2(2)=24.30,
\qquad
p=5.30\times10^{-6}.
\label{eq:context-completeness-chi-square}
\end{equation}

Because the audit allocated the same number of items to each band, the pooled count of 23 out of 90 is not an unweighted estimate of benchmark-wide prevalence. The Hard result of 17 out of 30, or 56.7\%, is a sample estimate at the audit stage. It is not a claim about context completeness after the later full Hard recheck, and the audit does not establish that every sampled case was automatically corrected. The concentration in the Hard band is expected because missing exhibits and detached vignettes deprive models of the evidence required by the released record. This is a data representation issue, not evidence that the underlying professional question is invalid.

\subsection{Context-Completeness Labeler Calibration}
\label{app:context-completeness-calibration}

The context-completeness labeler was calibrated after the manual audit, using the 90 audited items as reference labels and treating context incomplete as the positive class. It agreed with the manual labels on 88 of 90 items.

\begin{table}[t]
\centering
\small
\setlength{\tabcolsep}{5.0pt}
\renewcommand{\arraystretch}{1.08}
\begin{tabular}{@{}lrl@{}}
\toprule
\textbf{Metric}
& \textbf{Observed}
& \textbf{Wilson 95\% CI} \\
\midrule
Agreement & \(88/90=97.8\%\) & \([92.3\%,99.4\%]\) \\
Recall    & \(23/23=100.0\%\) & \([85.7\%,100.0\%]\) \\
Precision & \(23/25=92.0\%\) & \([75.0\%,97.8\%]\) \\
\bottomrule
\end{tabular}
\caption{Calibration of the context-completeness labeler against the stratified manual audit.}
\label{tab:context-completeness-calibration}
\end{table}

All 23 manually identified positive examples were detected, so the observed audit contains no missed positives. The recall interval nevertheless has a lower bound of 85.7\% because recall is estimated from only 23 positive examples. This lower bound reflects the limited positive sample size, not observed false negatives.

\subsection{Context-Complete Reasoning Track}
\label{app:context-complete-track}

The Full-Coverage Track contains 10{,}198 items and supports complete
coverage, difficulty construction, and the maintained leaderboard. The
Context-Complete Reasoning Track contains 7{,}625 items and is the primary
evidence base for financial reasoning from the supplied local record. It
preserves the single difficulty classification frozen on all
10{,}198 items rather than constructing a second difficulty scale. Thus,
the full Hard band contains 1{,}437 items and its Context-Complete Hard
subset contains 372 items without re-banding or band switching.

\subsubsection{Full Hard Context Recheck for Sensitivity Analysis}
\label{app:hard-context-recheck}

After labeler calibration, the lead curator manually rechecked every item assigned to the frozen Hard band. Together, the calibrated labeler and this full-band review established a separate context-completeness status for sensitivity analysis. The pass did not redefine or overwrite the frozen empirical band, and it should not be interpreted as evidence that every case identified in the earlier sample was corrected. Context-sensitive conclusions concerning Hard items are therefore recomputed on the subset marked context complete by this later assessment.

Of all 10{,}198 items, 7{,}625 form the Context-Complete Reasoning Track. The context-complete rate is 66.7\% for Public and 82.9\% for Held-out. Context completeness declines strongly with difficulty, reaching only 25.9\% for Hard items. The reasoning track is a question-level context-completeness filter, not a second difficulty taxonomy. All retained questions keep the frozen \texttt{difficulty} labels assigned on the Full-Coverage Track, and the filter neither recomputes the bands nor permits band switching. The 5{,}929 Easy, 1{,}324 Medium, and 372 Hard questions are the numbers from the original fixed bands that pass this filter. The 372 Context-Complete Hard items are distinct from the 188-item universal-failure core used in the shared-error analysis. Table~\ref{tab:context-complete-access-difficulty} cross-tabulates access partition and fixed difficulty within this track.

\begin{table}[t]
\centering
\small
\setlength{\tabcolsep}{3.2pt}
\renewcommand{\arraystretch}{1.08}
\begin{tabular}{@{}llrr@{}}
\toprule
\textbf{Dimension}
& \textbf{Category}
& \textbf{Full set}
& \textbf{Context-complete} \\
\midrule
& All questions & 10{,}198 & 7{,}625 \\
\midrule
\multirow{2}{*}{Access}
& Public   & 5{,}110 & 3{,}406 \\
& Held-out & 5{,}088 & 4{,}219 \\
\midrule
\multirow{3}{*}{Difficulty}
& Easy   & 6{,}578 & 5{,}929 \\
& Medium & 2{,}183 & 1{,}324 \\
& Hard   & 1{,}437 & 372 \\
\bottomrule
\end{tabular}
\caption{The Full-Coverage Track comprises all 10{,}198 questions. The Context-Complete Reasoning Track comprises the 7{,}625 questions retained by a question-level context-completeness filter. Each retained item keeps the frozen \texttt{difficulty} label assigned on the Full-Coverage Track, with no re-banding or band switching.}
\label{tab:context-complete-composition}
\end{table}

\begin{table}[t]
\centering
\small
\setlength{\tabcolsep}{5.0pt}
\renewcommand{\arraystretch}{1.08}
\begin{tabular}{@{}lrrr@{}}
\toprule
\textbf{Difficulty}
& \textbf{Public}
& \textbf{Held-out}
& \textbf{Total} \\
\midrule
Easy   & 2{,}714 & 3{,}215 & 5{,}929 \\
Medium & 554     & 770     & 1{,}324 \\
Hard   & 138     & 234     & 372 \\
\midrule
\textbf{Total} & \textbf{3{,}406} & \textbf{4{,}219} & \textbf{7{,}625} \\
\bottomrule
\end{tabular}
\caption{Access partition by fixed \texttt{difficulty} within the 7{,}625-item Context-Complete Reasoning Track. Counts use the original frozen difficulty labels from the Full-Coverage Track.}
\label{tab:context-complete-access-difficulty}
\end{table}

\begin{table*}[t]
\centering
\scriptsize
\setlength{\tabcolsep}{3.8pt}
\renewcommand{\arraystretch}{1.06}
\begin{tabularx}{0.98\textwidth}{@{}l*{6}{>{\raggedleft\arraybackslash}X}@{}}
\toprule
\textbf{Model}
& \textbf{CFA Level I}
& \textbf{CFA Level II}
& \textbf{CFA Level III}
& \textbf{FRM Part I}
& \textbf{FRM Part II}
& \textbf{All} \\
\midrule
\multicolumn{7}{@{}l}{\textit{Proprietary models}} \\
Gemini-3.1-Pro  & 98.11 & 93.94 & 90.53 & 95.48 & 91.02 & 95.96 \\
GPT-5.6-Sol     & 96.84 & 93.94 & 89.61 & 95.76 & 91.30 & 95.32 \\
GPT-5.6-Terra   & 94.51 & 90.66 & 86.84 & 95.69 & 89.23 & 93.29 \\
DeepSeek-V4-Pro & 94.73 & 91.05 & 87.07 & 93.53 & 88.67 & 93.01 \\
GPT-5.6-Luna    & 93.27 & 88.57 & 84.53 & 92.14 & 85.36 & 91.19 \\
Claude-Sonnet-5 & 91.97 & 86.58 & 85.68 & 91.79 & 85.36 & 90.24 \\
DeepSeek-R1     & 92.54 & 88.47 & 84.99 & 90.96 & 85.77 & 90.64 \\
Qwen3.7-Max     & 87.80 & 81.31 & 80.14 & 83.10 & 78.87 & 84.77 \\
GPT-5.5         & 88.30 & 81.21 & 78.29 & 86.93 & 82.87 & 86.02 \\
GPT-4o          & 82.68 & 73.66 & 71.59 & 74.83 & 70.86 & 78.26 \\
\addlinespace[1pt]
\multicolumn{7}{@{}l}{\textit{Open-weight reasoning models}} \\
GPT-OSS-120B    & 88.37 & 81.71 & 81.99 & 88.32 & 75.28 & 85.88 \\
GPT-OSS-20B     & 82.08 & 75.05 & 74.13 & 81.78 & 68.92 & 79.40 \\
\addlinespace[1pt]
\multicolumn{7}{@{}l}{\textit{Finance-specialized models}} \\
ODA-Fin-RL-8B   & 82.65 & 71.37 & 69.05 & 76.84 & 65.19 & 77.64 \\
Fin-o1-14B      & 78.75 & 68.99 & 65.36 & 74.13 & 64.78 & 74.50 \\
DianJin-R1-32B  & 76.42 & 67.00 & 66.28 & 71.84 & 67.13 & 72.85 \\
Hawkish-8B      & 60.83 & 52.19 & 53.81 & 53.41 & 45.99 & 56.49 \\
Fin-R1-7B       & 62.97 & 54.27 & 56.12 & 56.05 & 50.55 & 58.95 \\
\bottomrule
\end{tabularx}
\caption{Accuracy (\%) of the frozen panel of 17 models on the 7{,}625-item Context-Complete Reasoning Track by examination stage. DeepSeek-R1 is retained in the proprietary group, consistent with the main leaderboard.}
\label{tab:context-complete-leaderboard}
\end{table*}

All models in the frozen panel of 17 models score higher after the context
filter. Full-Coverage Track and Context-Complete Reasoning Track rankings
nevertheless remain nearly identical, with Spearman's \(\rho=0.988\) and
\(p=1.63\times10^{-13}\). The core model ranking and broad performance gap
between proprietary and finance-specialized models persist.

\subsection{Diagnostic Slice Definitions}
\label{app:diagnostic-slices}

We use two distinct diagnostic subsets for complementary analyses. The \emph{Context-Complete Hard subset} contains the frozen Hard items whose current records retain sufficient local context for independent resolution. The \emph{universal-failure core} contains the items for which every member of the frozen panel of 17 models returned a parseable but incorrect option. These subsets serve different purposes and are not interchangeable.

\begin{table}[t]
\centering
\small
\setlength{\tabcolsep}{4.0pt}
\renewcommand{\arraystretch}{1.08}
\begin{tabular}{@{}lrr@{}}
\toprule
\textbf{Slice}
& \textbf{\# Items}
& \textbf{Reference Share} \\
\midrule
Full-Coverage Track
& 10{,}198
& 100.0\% of full \\
Frozen Hard band
& 1{,}437
& 14.1\% of full \\
Context-Complete Hard
& 372
& 25.9\% of Hard \\
Universal-failure core
& 188
& 13.1\% of Hard \\
\bottomrule
\end{tabular}
\caption{Definitions and sizes of the principal diagnostic slices. The 372 Context-Complete Hard items and the 188 universal-failure items are distinct subsets of the 1{,}437-item frozen Hard band.}
\label{tab:diagnostic-slices}
\end{table}

\begin{table}[t]
\centering
\small
\setlength{\tabcolsep}{4.5pt}
\renewcommand{\arraystretch}{1.06}
\begin{tabular}{@{}llrr@{}}
\toprule
\textbf{Dimension}
& \textbf{Category}
& \textbf{\# Items}
& \textbf{Share} \\
\midrule
\multirow{2}{*}{Access}
& Public   & 118 & 62.8\% \\
& Held-out & 70  & 37.2\% \\
\midrule
\multirow{5}{*}{Stage}
& CFA Level II  & 79 & 42.0\% \\
& CFA Level I   & 46 & 24.5\% \\
& FRM Part I    & 24 & 12.8\% \\
& CFA Level III & 22 & 11.7\% \\
& FRM Part II   & 17 & 9.0\% \\
\bottomrule
\end{tabular}
\caption{Composition of the 188-item universal-failure core. Every member of the frozen panel of 17 models returned a parseable answer option on every item, so no item enters this set because of a parsing or truncation failure.}
\label{tab:universal-failure-composition}
\end{table}

\section{Empirical Difficulty and Statistical Details}
\label{app:difficulty}

\subsection{Constructing \texttt{difficulty}}

\paragraph{Frozen response matrix.}
After completion of the full two-stage expert reannotation, a frozen panel of 17 models answered each of the 10{,}198 Full-Coverage Track items. The panel contains 10 API-served models accessed through their respective providers and seven local models evaluated on H100 GPUs. DianJin used two H100 GPUs, whereas each of the other six local models used one. A valid answer is scored against the gold option. Parse failures and length-truncated outputs are scored as incorrect, leaving no missing cells in the \(17\times10{,}198\) matrix. The matrix and checksum are versioned so that the reported bands remain tied to the frozen evaluation version.

\paragraph{Group-balanced score.}
Let \(c_{m,i}\in\{0,1\}\) indicate whether model \(m\) answers item \(i\) correctly. We compute separate group means
\begin{equation}
\mu_{g,i}
=
\frac{1}{|\mathcal{M}_g|}
\sum_{m\in\mathcal{M}_g} c_{m,i},
\end{equation}
where \(g\in\{\mathrm{API},\mathrm{local}\}\). The adopted Rule C score weights the two means equally:
\begin{equation}
s_i
=
\frac{1}{2}
\left(
\frac{1}{10}
\sum_{m\in\mathcal{M}_{\mathrm{API}}}c_{m,i}
+
\frac{1}{7}
\sum_{m\in\mathcal{M}_{\mathrm{local}}}c_{m,i}
\right).
\label{eq:consensus}
\end{equation}
Equal weighting is used only to prevent the numerically larger API group from receiving mechanically greater influence. The grouping is not presented as a theoretical taxonomy of model capability.

The empirical band is
\begin{equation}
\operatorname{band}(i)=
\begin{cases}
\textsc{Easy},   & s_i \geq 2/3,\\
\textsc{Hard},   & s_i \leq 1/3,\\
\textsc{Medium}, & \text{otherwise}.
\end{cases}
\label{eq:band}
\end{equation}
These bands measure difficulty for the frozen model panel. They are not expert-assigned curriculum levels.

\begin{table}[t]
\centering
\small
\setlength{\tabcolsep}{5.0pt}
\renewcommand{\arraystretch}{1.08}
\begin{tabular}{@{}lrrl@{}}
\toprule
\textbf{Band}
& \textbf{\# Items}
& \textbf{Share}
& \textbf{Wilson 95\% CI} \\
\midrule
Easy   & 6{,}578 & 64.50\% & \([63.57\%,65.43\%]\) \\
Medium & 2{,}183 & 21.41\% & \([20.62\%,22.21\%]\) \\
Hard   & 1{,}437 & 14.09\% & \([13.43\%,14.78\%]\) \\
\bottomrule
\end{tabular}
\caption{Composition of \texttt{difficulty}. The counts are fixed properties of this frozen dataset version. The Wilson intervals describe uncertainty only under an item-sampling interpretation and do not express uncertainty about the reported counts.}
\label{tab:difficulty-distribution}
\end{table}

\subsection{Avoiding Circular Evaluation}
\label{app:circularity}

Per-model difficulty results use leave-one-model-out bands. When evaluating model \(m\), we recompute Equation~\ref{eq:consensus} using only the other 16 models, recompute the within-group denominators, and then apply Equation~\ref{eq:band}. Model \(m\) therefore does not help define the difficulty subset on which its performance is reported. The resulting per-model band sizes can differ slightly from the fixed global counts in Table~\ref{tab:difficulty-distribution}.

\subsection{Robustness of the Difficulty Partition}

\paragraph{Alternative panel weightings.}
Rule A is the flat mean across all 17 models. Rule B separates API-served, open-weight general-purpose, and open-weight finance-specialized models and weights the three group means equally. Rule C is the adopted API versus local grouping in Equation~\ref{eq:consensus}.

\begin{table}[t]
\centering
\scriptsize
\setlength{\tabcolsep}{3.0pt}
\renewcommand{\arraystretch}{1.06}
\begin{tabular}{@{}llrrr@{}}
\toprule
\textbf{Rule} & \textbf{Panel Weighting}
& \textbf{Easy} & \textbf{Medium} & \textbf{Hard} \\
\midrule
A & Flat mean over 17 models
& 6{,}826 & 2{,}025 & 1{,}347 \\
B & Three group means, equal weight
& 6{,}757 & 1{,}714 & 1{,}727 \\
\textbf{C} & \textbf{API and local means, equal weight}
& \textbf{6{,}578} & \textbf{2{,}183} & \textbf{1{,}437} \\
\bottomrule
\end{tabular}
\caption{Band sizes under alternative panel-weighting rules. Rule C defines \texttt{difficulty}.}
\label{tab:difficulty-rules}
\end{table}

Rule C and Rule A assign the same band to 95.8\% of items, and their continuous scores have Spearman correlation \(\rho=0.998\). Rule C and Rule B agree on 90.8\% of band assignments, with \(\rho=0.991\) between their continuous scores.

\paragraph{Panel stability.}
We formed 200 random split halves of the 17 models, with eight models in one half and nine in the other, and compared the resulting item orderings. The mean split-half Spearman correlation was \(0.821\), with standard deviation \(0.024\) and range \(0.737\) to \(0.849\). The Spearman--Brown corrected reliability was \(0.902\). Across leave-one-model-out reconstructions, the worst correlation between a 16-model score and the full-panel score was \(0.983\). No single model therefore determines the empirical ordering.

\subsection{Relation to Examination Stage}

Empirical difficulty and examination stage are retained as separate variables. The association between the three empirical bands and the five stages is modest, with Cram\'er's \(V=0.174\). The association with examination family alone is smaller, with \(V=0.056\).

\begin{table}[t]
\centering
\small
\setlength{\tabcolsep}{4.2pt}
\renewcommand{\arraystretch}{1.06}
\begin{tabular}{@{}lrrrr@{}}
\toprule
\textbf{Stage} & \textbf{Items}
& \textbf{Easy} & \textbf{Medium} & \textbf{Hard} \\
\midrule
CFA Level I   & 4{,}763 & 72.9\% & 17.3\% & 9.8\% \\
CFA Level II  & 2{,}215 & 46.5\% & 28.0\% & 25.5\% \\
CFA Level III & 653     & 57.1\% & 23.0\% & 19.9\% \\
FRM Part I    & 1{,}652 & 71.7\% & 19.6\% & 8.7\% \\
FRM Part II   & 915     & 56.4\% & 29.1\% & 14.5\% \\
\bottomrule
\end{tabular}
\caption{Frozen empirical difficulty composition within each examination stage.}
\label{tab:difficulty-by-stage}
\end{table}

CFA Level II has the largest Hard share, whereas CFA Level III has a lower Hard share despite its later curriculum position. Empirical model difficulty therefore does not reproduce a common ordinal progression across the CFA and FRM stages.

\begin{figure*}[t]
    \centering
    \begin{minipage}[t]{0.48\textwidth}
        \centering
        \includegraphics[width=\linewidth]
        {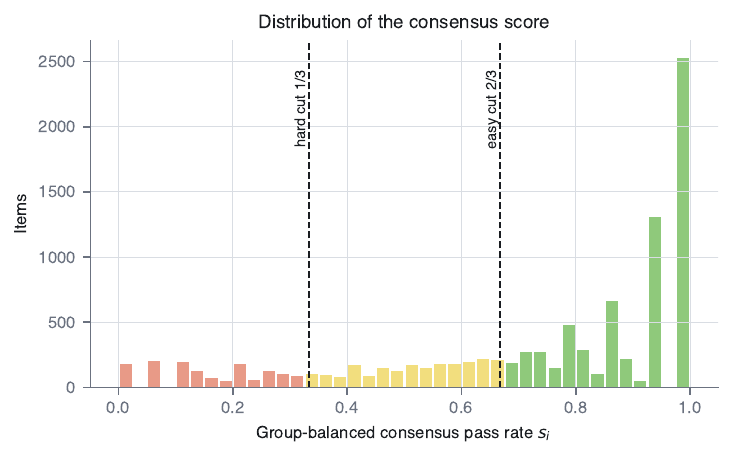}

        \smallskip
        \small (a) Consensus distribution and empirical-band thresholds.
    \end{minipage}
    \hfill
    \begin{minipage}[t]{0.48\textwidth}
        \centering
        \includegraphics[width=\linewidth]
        {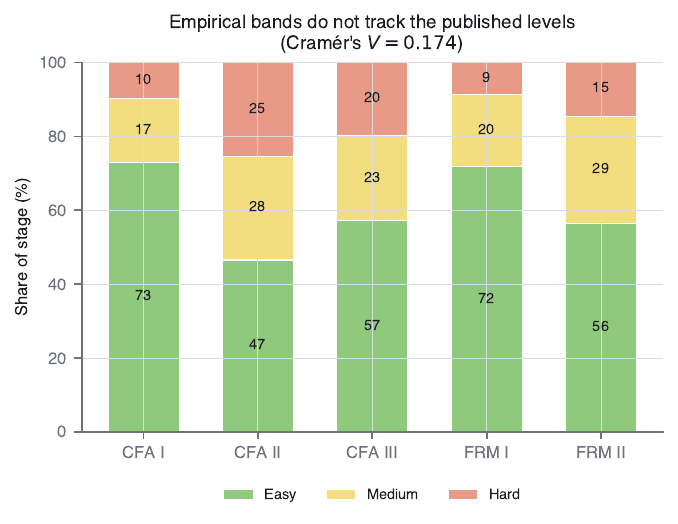}

        \smallskip
        \small (b) Empirical-band composition by examination stage.
    \end{minipage}
    \caption{Construction and validation of the empirical difficulty band on all 10{,}198 Full-Coverage Track items. The group-balanced consensus score comes from the frozen panel of 17 models, with Hard at $s_i\leq1/3$ and Easy at $s_i\geq2/3$. The stage association is modest (Cram\'er's $V=0.174$), so empirical difficulty is distinct from CFA and FRM curriculum stage.}
    \label{fig:difficulty-construction-validation}
\end{figure*}

Figure~\ref{fig:difficulty-construction-validation} visualizes both the consensus thresholds and their modest association with curriculum stage.

\subsection{Statistical Procedures}
\label{app:tests}

\paragraph{Wilson intervals and fixed counts.}
For an observed proportion \(\hat{p}=x/n\) and \(z=1.96\), the Wilson interval is
\begin{equation}
\frac{
\hat{p}+\frac{z^2}{2n}
\ \pm\
z\sqrt{
\frac{\hat{p}(1-\hat{p})}{n}
+
\frac{z^2}{4n^2}
}
}{
1+\frac{z^2}{n}
}.
\label{eq:wilson}
\end{equation}
For the context-completeness audit, these intervals quantify uncertainty from the finite audited sample. For the Easy, Medium, and Hard shares of \texttt{difficulty}, they describe uncertainty only under a hypothetical item-sampling interpretation. The observed counts of 6{,}578, 2{,}183, and 1{,}437 are fixed properties of the frozen version and have no sampling uncertainty within that version.

\paragraph{Curation and context-completeness tests.}
The curation-yield comparison uses a \(2\times2\) Pearson chi-square test of examination family by final retention status. The context-completeness audit uses a Pearson chi-square test of the three frozen bands by binary context-completeness status. The reported statistics are \(\chi^2(1)=92.47\), \(p=6.83\times10^{-22}\), for curation yield and \(\chi^2(2)=24.30\), \(p=5.30\times10^{-6}\), for context completeness.

\paragraph{Partition-direction tests.}
The public versus held-out analysis treats each model as one paired directional observation. Under a null probability of \(1/2\) for either direction, a unanimous result across 17 models has the exact two-sided sign-test probability given in Equation~\ref{eq:partition-sign-test}. On the Full-Coverage Track, the unanimous direction is lower public accuracy. On the Context-Complete Reasoning Track, the unanimous direction is higher public accuracy.

\paragraph{Accuracy relative to chance.}
CFA questions have three options and FRM questions have four. A mixed CFA and FRM subset therefore has item-specific random-guess probabilities of \(1/3\) and \(1/4\). The number of correct random responses follows a Poisson-binomial distribution rather than a single binomial distribution. We compute the exact one-sided tail probability by dynamic programming over the item-level probabilities and control the false discovery rate across the 17 models at \(q<0.05\).

\paragraph{Paired intervention comparisons.}
Methods evaluated on matched items are compared using the exact McNemar test on discordant response pairs. The two discordant counts correspond to baseline incorrect followed by method correct and baseline correct followed by method incorrect. These counts also determine the rescue rate, harm rate, and net repair.

\paragraph{Concentration of shared errors.}
Let \(\mathcal{U}\) denote the \(188\) items answered incorrectly by all \(17\) models under the frozen direct baseline. Every item in \(\mathcal{U}\) belongs to the global Hard band because its full-panel consensus score is zero. The set contains \(118\) public and \(70\) held-out items. Its stage composition is \(79\) CFA Level II, \(46\) CFA Level I, \(24\) FRM Part I, \(22\) CFA Level III, and \(17\) FRM Part II items. All \(17\) models returned a parseable option letter for every item in \(\mathcal{U}\), so membership in this set is not caused by answer-extraction or truncation failures.

For each item \(i\in\mathcal{U}\), distractor concentration is
\begin{equation}
C_i=\frac{\max_j n_{ij}}{17},
\label{eq:error-concentration}
\end{equation}
where \(n_{ij}\) is the number of models selecting distractor \(j\). The item-specific null distributes the \(17\) wrong answers uniformly over the available distractors, two for CFA items and three for FRM items. Under this null, the exact expected modal share is \(0.5982\) for a CFA item and \(0.4539\) for an FRM item. Because 147 of the 188 items are CFA questions, the median item-specific null expectation for \(\mathcal{U}\) is \(0.5982\). We compare observed and null concentration with a paired Wilcoxon signed-rank test and report the median paired excess together with the share of items exceeding their own null expectation and its Wilson 95\% confidence interval.

\paragraph{Association, reliability, and multiplicity.}
Cram\'er's \(V\) measures associations between categorical variables. Spearman's \(\rho\) measures agreement between continuous difficulty scores, split-half scores, and leave-one-model-out scores. Split-half reliability is adjusted with the Spearman--Brown formula. Effect sizes accompany significance tests, and Benjamini--Hochberg correction is applied within each predeclared family of model comparisons.

\paragraph{Scope of inference.}
The 17 models constitute a deliberately broad evaluation panel rather than a random sample from all possible language models. Statistical intervals therefore characterize item-level uncertainty for this panel and do not establish population-level performance for arbitrary future models. Some CFA questions share a vignette, which introduces dependence among sibling items. Item-level tests are retained for comparability with prior benchmark evaluations, but very small \(p\)-values are interpreted together with effect sizes, context-completeness audits, and sensitivity analyses.

\section{RQ1 Diagnostic Analyses}
\label{app:rq1-diagnostics}

This appendix expands the benchmark diagnostics summarized in RQ1. We
distinguish the Full-Coverage Track, the frozen Hard band, the Context-
Complete Reasoning Track, the Context-Complete Hard subset, the
universal-failure core, and the intersection of the last two. The
Context-Complete Reasoning Track controls for representation-level context loss,
whereas the universal-failure core isolates items missed by the fixed
17 model panel. All 188 core items belong to the frozen Hard band, and
all baseline models returned parseable option letters on every core
item. These slices serve complementary purposes and are not
interchangeable.

\begin{table}[t]
\centering
\scriptsize
\setlength{\tabcolsep}{2.8pt}
\renewcommand{\arraystretch}{1.06}
\begin{tabularx}{\columnwidth}{@{}>{\raggedright\arraybackslash}Xrr@{}}
\toprule
\textbf{Diagnostic slice}
& \textbf{\# Items}
& \textbf{Reference share} \\
\midrule
Full-Coverage Track
& 10{,}198
& 100.0\% of full \\
Frozen Hard band
& 1{,}437
& 14.1\% of full \\
Context-Complete Reasoning Track
& 7{,}625
& 74.8\% of full \\
Context-Complete Hard
& 372
& 25.9\% of Hard \\
Universal-failure core
& 188
& 13.1\% of Hard \\
Their intersection
& 47
& 25.0\% of \(\mathcal{U}\) \\
\bottomrule
\end{tabularx}
\caption{
Definitions and sizes of the principal RQ1 diagnostic slices.
The 47-item intersection contains questions that are both
Context-Complete Hard and missed by the frozen panel of 17 models.
}
\label{tab:rq1-slice-definitions}
\end{table}

\subsection{Chance Baselines and Subject Normalization}
\label{app:chance-normalization}

For item \(i\), let \(K_i\) denote the number of answer options.
Uniform random guessing gives probability \(1/K_i\), with \(K_i=3\)
for CFA and \(K_i=4\) for FRM. For a mixed subset
\(\mathcal{S}\), its item-weighted chance rate is

\begin{equation}
b_{\mathcal{S}}
=
\frac{1}{|\mathcal{S}|}
\sum_{i\in\mathcal{S}}\frac{1}{K_i}.
\label{eq:item-weighted-chance}
\end{equation}

Let \(a_{m,\mathcal{S}}\in[0,1]\) be model \(m\)'s observed accuracy on
\(\mathcal{S}\). We define chance-normalized accuracy as

\begin{equation}
a^{*}_{m,\mathcal{S}}
=
100
\frac{a_{m,\mathcal{S}}-b_{\mathcal{S}}}
     {1-b_{\mathcal{S}}}.
\label{eq:chance-normalized-accuracy}
\end{equation}

Figure~\ref{fig:subject-radar} uses the same transformation within each
subject. Subject-level analysis is restricted to the 5{,}088 held-out
items with complete curriculum metadata and retains the 15 subjects
containing at least 150 items. The 372-item Context-Complete Hard subset
contains 241 CFA and 131 FRM questions, yielding
\(b_{\mathcal{H}_{\mathrm{cc}}}=0.3040\).

\subsection{Context-Complete Hard Performance}
\label{app:context-complete-hard-results}

Let \(d_i\) denote the frozen empirical difficulty label and
\(\gamma_i\in\{0,1\}\) the context-completeness flag. The
Context-Complete Hard subset is

{\small
\begin{equation}
\mathcal{H}_{\mathrm{cc}}
=
\left\{
i:
d_i=\textsc{Hard}
\land
\gamma_i=1
\right\},
\qquad
|\mathcal{H}_{\mathrm{cc}}|=372.
\label{eq:context-complete-hard}
\end{equation}
}

Filtering is applied after the difficulty freeze. It does not recompute
the empirical bands or permit an item to change bands.

The following table reports the frozen panel of 17 models on the 372-item
Context-Complete Hard subset. It is distinct from the 188-item
universal-failure core, for which every panel member is incorrect by
definition.

\begin{table}[t]
\centering
\scriptsize
\setlength{\tabcolsep}{3.6pt}
\renewcommand{\arraystretch}{1.06}
\begin{tabular}{@{}lrrr@{}}
\toprule
\textbf{Model}
& \textbf{Acc.}
& \textbf{Norm. Acc.}
& \textbf{95\% CI} \\
\midrule
\multicolumn{4}{@{}l}{\textit{Proprietary models}} \\[1pt]
Gemini-3.1-Pro  & \textbf{54.57} & \(+34.73\) & [49.5, 59.6] \\
GPT-5.6-Sol     & 39.52 & \(+13.10\) & [34.7, 44.6] \\
GPT-5.6-Terra   & 31.18 & \(+1.13\)  & [26.7, 36.1] \\
DeepSeek-V4-Pro & 26.08 & \(-6.21\)  & [21.9, 30.8] \\
Claude-Sonnet-5 & 21.24 & \(-13.16\) & [17.4, 25.7] \\
GPT-5.5          & 21.24 & \(-13.16\) & [17.4, 25.7] \\
GPT-5.6-Luna     & 18.01 & \(-17.80\) & [14.4, 22.2] \\
Qwen3.7-Max      & 14.78 & \(-22.43\) & [11.5, 18.8] \\
DeepSeek-R1      & 14.25 & \(-23.21\) & [11.1, 18.2] \\
GPT-4o           & 12.90 & \(-25.14\) & [9.9, 16.7] \\
\addlinespace[1pt]
\multicolumn{4}{@{}l}{\textit{Open-weight reasoning models}} \\[1pt]
GPT-OSS-20B      & 7.80 & \(-32.48\) & [5.5, 11.0] \\
GPT-OSS-120B     & 5.65 & \(-35.56\) & [3.7, 8.5] \\
\addlinespace[1pt]
\multicolumn{4}{@{}l}{\textit{Finance-specialized models}} \\[1pt]
Hawkish-8B       & 12.37 & \(-25.91\) & [9.4, 16.1] \\
Fin-o1-14B       & 12.10 & \(-26.30\) & [9.2, 15.8] \\
Fin-R1-7B        & 9.95 & \(-29.39\) & [7.3, 13.4] \\
ODA-Fin-RL-8B    & 6.99 & \(-33.63\) & [4.8, 10.0] \\
DianJin-R1-32B   & 5.91 & \(-35.18\) & [3.9, 8.8] \\
\bottomrule
\end{tabular}
\caption{
Performance of the frozen panel of 17 models on the 372-item
Context-Complete Hard subset, which is distinct from the 188-item
universal-failure core.
\textbf{Norm. Acc.} applies
Equation~\ref{eq:chance-normalized-accuracy} using the subset's
30.40\% item-weighted chance rate. Intervals are Wilson 95\%
confidence intervals. Only the first three models have point estimates
above the weighted chance baseline.
}
\label{tab:context-complete-hard-results}
\end{table}

\subsection{Stage Enrichment and Item Format}
\label{app:hard-slice-composition}

Let \(\mathcal{D}\) denote the Full-Coverage Track and
\(\mathcal{D}_s\) the items belonging to examination stage \(s\). We
measure stage enrichment as the stage share in
\(\mathcal{H}_{\mathrm{cc}}\) divided by its share in the full
benchmark:

\begin{equation}
E_s
=
\frac{
|\mathcal{H}_{\mathrm{cc}}\cap\mathcal{D}_s|
/
|\mathcal{H}_{\mathrm{cc}}|
}{
|\mathcal{D}_s|/|\mathcal{D}|
}.
\label{eq:stage-enrichment}
\end{equation}

\begin{table}[t]
\centering
\small
\setlength{\tabcolsep}{5.0pt}
\renewcommand{\arraystretch}{1.06}
\begin{tabular}{@{}lr@{}}
\toprule
\textbf{Examination stage}
& \textbf{Enrichment \(E_s\)} \\
\midrule
FRM Part II   & 1.98 \\
CFA Level III & 1.93 \\
FRM Part I    & 1.08 \\
CFA Level I   & 0.78 \\
CFA Level II  & 0.73 \\
\bottomrule
\end{tabular}
\caption{
Stage enrichment in the Context-Complete Hard subset. Values above
one indicate overrepresentation relative to the Full-Coverage Track.
}
\label{tab:context-hard-stage-enrichment}
\end{table}

The context filter also changes the observed item-format mixture. The
two categories in Table~\ref{tab:hard-item-format} are diagnostic and
not exhaustive.

\begin{table}[t]
\centering
\footnotesize
\setlength{\tabcolsep}{2.5pt}
\renewcommand{\arraystretch}{1.06}
\begin{tabular}{@{}lcc@{}}
\toprule
\textbf{Slice}
& \textbf{\shortstack{Judgment /\\discrimination}}
& \textbf{\shortstack{Numeric-option\\calculation}} \\
\midrule
Full-Coverage Track
& 44\% & 20\% \\
Context-Complete Hard
& 53\% & 6\% \\
Context-incomplete Hard
& 21\% & 46\% \\
\bottomrule
\end{tabular}
\caption{
Shares of two diagnostic item formats. Missing contextual material is
concentrated among calculation questions, whereas the
Context-Complete Hard subset contains more judgment and discrimination
questions. The remaining questions use other formats.
}
\label{tab:hard-item-format}
\end{table}

Across the item-format groups used in this audit, panel-averaged
accuracy within the Context-Complete Hard subset lies in a comparatively
narrow 17--22\% range. The remaining difficulty is therefore not
explained by one dominant format alone.

\subsection{Exact Distractor-Concentration Null}
\label{app:hard-distractor-concentration}

For item \(i\), let \(r_i\) be the number of model predictions assigned
to a non-gold option and \(n_{ij}\) the number assigned to distractor
\(j\). The modal distractor concentration is

\begin{equation}
C_i
=
\frac{\max_j n_{ij}}{r_i}.
\label{eq:hard-concentration}
\end{equation}

The statistic is defined for 369 Context-Complete Hard items with at
least one recorded distractor vote. The appropriate null is not
\(1/K_i\). Conditional on being wrong, a CFA prediction can fall on two
distractors and an FRM prediction on three. Let \(D_i=K_i-1\). Under
independent uniform distractor selection,

{\small
\begin{equation}
(N_{i1},\ldots,N_{iD_i})
\sim
\operatorname{Multinomial}
\left(
r_i,
\frac{1}{D_i},\ldots,\frac{1}{D_i}
\right).
\label{eq:distractor-multinomial}
\end{equation}
}

The exact expected modal share is

\begin{equation}
\mu_i
=
\sum_{\substack{n_1+\cdots+n_{D_i}=r_i\\n_j\geq0}}
\frac{\max_j n_j}{r_i}
\frac{r_i!}{\prod_{j=1}^{D_i}n_j!}
\left(\frac{1}{D_i}\right)^{r_i}.
\label{eq:exact-modal-null}
\end{equation}

\begin{table}[t]
\centering
\small
\setlength{\tabcolsep}{4.0pt}
\renewcommand{\arraystretch}{1.06}
\begin{tabular}{@{}lr@{}}
\toprule
\textbf{Statistic}
& \textbf{Observed} \\
\midrule
Items with defined concentration
& 369 \\
Median \(C_i\)
& 0.923 \\
Median exact null \(\mu_i\)
& 0.605 \\
Items with \(C_i>\mu_i\)
& 346/369 (93.8\%) \\
All erroneous votes on one distractor
& 148/369 (40.1\%) \\
One-sided exact sign-test \(p\)
& \(1.9\times10^{-75}\) \\
\bottomrule
\end{tabular}
\caption{
Distractor concentration on Context-Complete Hard items. The null
conditions on the number of erroneous votes and allocates them
uniformly over the non-gold options. The sign test evaluates the
directional hypothesis \(C_i>\mu_i\).
}
\label{tab:hard-distractor-concentration}
\end{table}

For completeness, the exact directional sign-test probability is

\begin{equation}
p_{\mathrm{sign}}
=
\sum_{k=346}^{369}
\binom{369}{k}2^{-369}.
\label{eq:concentration-sign-test}
\end{equation}

The observed concentration is inconsistent with independent uniform
distractor selection for this fixed model panel. It does not establish
that the models share one internal causal mechanism. Professional
examination distractors are intentionally designed around plausible
mistakes and may themselves induce cross-model convergence.

\subsection{Universal-Failure Core}
\label{app:universal-failure-core}

Let \(c_{m,i}\in\{0,1\}\) indicate whether model \(m\) answers item
\(i\) correctly. The universal-failure core is

\begin{equation}
\mathcal{U}
=
\left\{
i:
\sum_{m=1}^{17}c_{m,i}=0
\right\},
\qquad
|\mathcal{U}|=188.
\label{eq:universal-failure-core}
\end{equation}

Every item in \(\mathcal{U}\) has a full-panel consensus score of zero
and therefore belongs to the frozen Hard band.

\begin{table}[t]
\centering
\small
\setlength{\tabcolsep}{4.2pt}
\renewcommand{\arraystretch}{1.06}
\begin{tabular}{@{}llrr@{}}
\toprule
\textbf{Dimension}
& \textbf{Category}
& \textbf{\# Items}
& \textbf{Share} \\
\midrule
\multirow{2}{*}{Access}
& Public   & 118 & 62.8\% \\
& Held-out & 70  & 37.2\% \\
\midrule
\multirow{5}{*}{Stage}
& CFA Level II  & 79 & 42.0\% \\
& CFA Level I   & 46 & 24.5\% \\
& FRM Part I    & 24 & 12.8\% \\
& CFA Level III & 22 & 11.7\% \\
& FRM Part II   & 17 & 9.0\% \\
\bottomrule
\end{tabular}
\caption{
Composition of the 188-item universal-failure core across access
partitions and examination stages.
}
\label{tab:universal-failure-stage-detail}
\end{table}

\begin{table}[t]
\centering
\small
\setlength{\tabcolsep}{4.2pt}
\renewcommand{\arraystretch}{1.06}
\begin{tabular}{@{}lrr@{}}
\toprule
\textbf{Universal-failure subset}
& \textbf{\# Items}
& \textbf{Share} \\
\midrule
Context incomplete
& 141 & 75.0\% \\
Context complete
& 47 & 25.0\% \\
\midrule
All 17 choose the same wrong option
& 41 & 21.8\% \\
\bottomrule
\end{tabular}
\caption{
Context completeness and unanimous distractor selection within the
universal-failure core. The 47 context-complete items are exactly
\(\mathcal{U}\cap\mathcal{H}_{\mathrm{cc}}\).
}
\label{tab:universal-failure-context}
\end{table}

Across the 188 items, the median modal distractor share is 0.806.
Because 147 items are CFA questions and 41 are FRM questions, the
median item-specific uniform null is 0.5982. Thus, even within the
strictest failure slice, errors frequently concentrate on a small
number of professionally plausible distractors.

\subsection{Answer-Position Diagnostics}
\label{app:option-position}

The difficult slices exhibit non-uniform gold-label and prediction
distributions. The following analyses are descriptive and do not imply
that answer position accounts for every error.

\paragraph{Universal-failure core.}
Option A is the gold label for 44.7\% of universal-failure items,
compared with 29.9\% of the Full-Coverage Track. Across the model
predictions on this slice, option B is selected most often, accounting
for 39.0\% of predictions. This mismatch indicates that part of the
universal-failure rate may be associated with answer-position
preference rather than financial content alone.

\paragraph{Context-Complete Hard subset.}
Table~\ref{tab:context-hard-position} compares the gold-position
distribution with the aggregate distribution of model selections.

\begin{table}[t]
\centering
\small
\setlength{\tabcolsep}{4.5pt}
\renewcommand{\arraystretch}{1.06}
\begin{tabular}{@{}lrrr@{}}
\toprule
\textbf{Position}
& \textbf{Gold}
& \textbf{Selected}
& \textbf{Gap} \\
\midrule
A & 35.2\% & 27.5\% & \(-7.7\) \\
B & 31.7\% & 29.6\% & \(-2.1\) \\
C & 25.0\% & 34.1\% & \(+9.1\) \\
D & 8.1\%  & 8.8\%  & \(+0.7\) \\
\bottomrule
\end{tabular}
\caption{
Gold-position and aggregate prediction-position distributions on the
372-item Context-Complete Hard subset. Gaps are percentage points.
Position D occurs only for FRM items. Its displayed shares are the
residual values implied by the reported A--C percentages and may
reflect rounding.
}
\label{tab:context-hard-position}
\end{table}

The direction differs from the universal-failure core. Models
under-select A and over-select C on Context-Complete Hard items,
whereas the universal-failure core overrepresents A among gold labels
and B among predictions. Answer-position effects should therefore be
treated as slice-specific confounds rather than as one fixed
benchmark-wide bias.

\section{Knowledge Intervention Details}
\label{app:intervention-details}

This section provides the formal definitions and implementation details omitted from Section~\ref{sec:evaluation-framework}. The presentation separates the externally transferred Function-RAG baseline from our FunctionGraph-RAG, verifier, and selective router.

\subsection{Notation and Matched Conditions}
\label{app:intervention-notation}

Let
\[
\mathcal{D}=\{(x_i,y_i)\}_{i=1}^{N}
\]
denote an evaluation set, where $x_i$ contains the question and answer options and $y_i$ is the gold option. For a branch
$b\in\{\mathrm{dir},\mathrm{func},\mathrm{fg},\mathrm{fgv}\}$, let
$o_i^b$ denote the complete model output, $\hat{y}_i^b$ the extracted option, and

\begin{equation}
c_i^b = \mathbb{I}\!\left[\hat{y}_i^b=y_i\right]
\label{eq:branch-correctness}
\end{equation}

its correctness indicator. Unparseable and length-truncated outputs have $c_i^b=0$. All matched comparisons use the same item identifiers, backbone, decoding budget, and answer-extraction procedure.

\subsection{Chain-Specific Function-RAG Baselines}
\label{app:function-rag-details}

We transfer the strongest corresponding Function-RAG configurations
from FinanceReasoning~\citep{tang-etal-2025-financereasoning}.
The two reasoning chains use different first-stage retrievers.

\paragraph{GPT-4o with PoT.}
An LLM-generated query retrieves the frozen Contriever top-30
functions. The transferred relevance judge retains zero to three
functions, whose signatures, descriptions, and implementations are
serialized with the question.

\paragraph{DeepSeek-R1 with CoT.}
An LLM-generated query retrieves the BM25 top-30 functions. The
transferred instructed-retrieval and relevance-judging stage then
selects the functions supplied to the CoT backbone.

Both baselines are transferred without adaptation. Neither uses
FinExam-10K held-out questions, answers, rationales, correctness
signals, or model outputs for training or model selection.

\subsection{FunctionGraph-RAG}
\label{app:functiongraph-rag-details}

FunctionGraph-RAG follows the Function-RAG formulation but is not an
exact reproduction. The GPT-4o PoT and DeepSeek-R1 CoT results use
separate frozen selectors rather than two inference modes of one
selector. Table~\ref{tab:functiongraph-selector-specs} records the full
chain-specific specifications.

\begin{table*}[t]
\centering
\scriptsize
\setlength{\tabcolsep}{4pt}
\renewcommand{\arraystretch}{1.12}
\begin{tabularx}{\textwidth}{@{}lXX@{}}
\toprule
\textbf{Component} & \textbf{GPT-4o PoT, Table~\ref{tab:rq2-full}} & \textbf{DeepSeek-R1 CoT, published R1 variants} \\
\midrule
Initial candidates
& Frozen Contriever top 30
& BM25 top 30 \\
Graph edges
& Two relation types. Shared \texttt{article\_title} edges and directed Contriever nearest neighbour edges with $k=4$ are unioned and deduplicated
& One relation type. Two functions are adjacent when their sets of normalized input and output quantity names intersect \\
Graph construction
& Static graph content built only from the function corpus. A FinExam-10K identifier manifest is read only to reject identifier overlap
& Function-quantity graph with the relation defined above \\
Expansion and cap
& One hop, with no candidate cap beyond collecting the one-hop expansion
& Up to two hops with a total cap of 80 candidates. Expansion stops as soon as the cap is reached \\
Selector
& Frozen linear selector with exactly four features: reciprocal retrieval rank, reciprocal graph depth, edge weight, and degree scale
& Learned linear reranker with exactly 56 features \\
Returned functions
& At most three
& Top ten \\
Training labels
& Pairwise ranking on 511 reachable direct \texttt{function\_id} relevance labels from FinanceReasoning Easy and Medium. Another 241 labels are excluded because the labeled function is absent from the candidate pool
& Pairwise ranking on 890 reachable direct \texttt{function\_id} relevance labels from FinanceReasoning Easy, Medium, and Hard \\
Split
& 403 train, 66 validation, 42 test
& 678 train, 106 development, 106 test \\
\bottomrule
\end{tabularx}
\caption{Chain-specific FunctionGraph-RAG selector specifications. The two selectors use different candidate generators, graph relations, expansion rules, ranking features, and output budgets.}
\label{tab:functiongraph-selector-specs}
\end{table*}

Both selectors use only FinanceReasoning labels for fitting and model
selection. Neither path uses FinExam-10K content, answers, rationales,
correctness, or model outputs. The PoT path reads only an identifier
manifest and fails if a training identifier overlaps FinExam-10K. The
selectors are frozen and transferred to all 10{,}198 Full-Coverage Track items
without adaptation. They are also distinct from the Function-RAG LLM
relevance judge. Selected functions and their induced relations are
serialized with the question and supplied to the corresponding reasoning
backbone.

\paragraph{Selector features and ranking objectives.}
For PoT, let $r_f$ be the zero based retrieval rank, $d_f$ the graph
depth, $e_f$ the edge weight, and $g_f$ the graph degree. The four feature
coordinates are
\begin{equation}
\phi_{\mathrm{PoT}}(f)
=
\left[
\frac{1}{r_f+1},
\frac{1}{d_f+1},
e_f,
\frac{g_f}{g_f+1}
\right].
\label{eq:pot-selector-features}
\end{equation}
The implementation constructs both directions of every relevant and
irrelevant pair and fits a linear logistic score for 200 full batch
updates. Candidates are ranked by the sigmoid of the linear score. The
frozen threshold is zero, so it does not remove a candidate, and the
first three ranked functions are returned. The graph expansion code uses
$e_f=1$ unless a candidate already supplies a weight.

For CoT, define $R(r)=1/r$ for a positive one based rank and zero
otherwise. Define $I_k(r)=\mathbb{I}[0<r\leq k]$ and
$O(A,B)=|A\cap B|/|A|$ when $A$ is nonempty, with zero otherwise. The 56
coordinates comprise a bias, normalized BM25 score and log score, BM25
rank features $R(r)$ and $I_k(r)$ for
$k\in\{1,3,5,10,20,40\}$, and a missing rank indicator. They also include
expanded rank features $R(r)$ and $I_k(r)$ for
$k\in\{1,3,5,10,40,80\}$, graph distance indicators for zero, one, two,
and missing distance, and graph reason count and reason type indicators.
Input and quantity features are clipped counts, zero input and generic
output indicators, and clipped mean and maximum quantity degrees. The
remaining lexical coordinates use $O$ between query tokens and the
function title, name, documentation, source, output, and inputs.
The code assigns 13 target or known value coordinates to zero and
duplicates the title, name, documentation, and source overlap coordinates.
Thus, the frozen vector has 56 positions but fewer distinct active
signals. It ranks by
\begin{equation}
s_{\mathrm{CoT}}(f)=\mathbf{w}^{\top}\phi_{\mathrm{CoT}}(f),
\label{eq:cot-selector-score}
\end{equation}
with BM25 rank and expansion order as tie breakers. Pairwise margin
training uses hard negatives. Development selection orders configurations
by MRR, hit at 10, hit at 3, and then fewer epochs. The final model is
fitted on train and development examples and returns the top ten.

An additional exploratory FunctionGraph-RAG top-10, no-judge
configuration does not correspond to either reported chain above. On DeepSeek-R1 CoT it
reaches 79.22\%, which is 0.53 points below Direct. Function-RAG BM25 and
LLM-judge configurations are external baselines rather than graph
variants.

\subsection{Post-Generation Verification}
\label{app:verifier-details}

For the GPT-4o PoT experiments, both verifier configurations trigger
only when Function-RAG and FunctionGraph-RAG disagree. The specified
verifier receives only the two answer letters. The main informed
verifier receives the item, answer options, both programs or traces,
and both execution results:
\begin{equation}
v_i
=
V_{\psi}\!\left(
x_i,\mathcal{O}_i,o_i^{\mathrm{func}},e_i^{\mathrm{func}},
o_i^{\mathrm{fg}},e_i^{\mathrm{fg}}
\right).
\label{eq:verifier}
\end{equation}
The informed verifier then selects the final answer, while agreement
passes through unchanged:
\begin{equation}
\hat{y}_i^{\mathrm{fgv}}
=
\begin{cases}
\hat{y}_i^{\mathrm{fg}},
& \hat{y}_i^{\mathrm{func}}=\hat{y}_i^{\mathrm{fg}},\\
T_{\mathrm{ver}}(v_i),
& \hat{y}_i^{\mathrm{func}}\neq\hat{y}_i^{\mathrm{fg}}.
\end{cases}
\label{eq:verified-output}
\end{equation}
The informed row reaches 68.92\%, or 0.45 points below Direct, on the
Full-Coverage Track. The verifier is post-generative and bounded. It does
not initiate a new retrieval trajectory or perform unrestricted
external search.

\subsection{Matched Transition Metrics}
\label{app:transition-metrics}

For an intervention branch $b$, we partition matched items into four
transition types:
\begin{align}
R_b
&=
\sum_{i=1}^{N}
\mathbb{I}
\!\left[
c_i^{\mathrm{dir}}=0
\land
c_i^{b}=1
\right],
\label{eq:rescue-count}
\\
H_b
&=
\sum_{i=1}^{N}
\mathbb{I}
\!\left[
c_i^{\mathrm{dir}}=1
\land
c_i^{b}=0
\right],
\label{eq:harm-count}
\\
S_b^{+}
&=
\sum_{i=1}^{N}
\mathbb{I}
\!\left[
c_i^{\mathrm{dir}}=1
\land
c_i^{b}=1
\right],
\label{eq:stable-correct}
\\
S_b^{-}
&=
\sum_{i=1}^{N}
\mathbb{I}
\!\left[
c_i^{\mathrm{dir}}=0
\land
c_i^{b}=0
\right].
\label{eq:persistent-wrong}
\end{align}
These counts satisfy
\[
R_b+H_b+S_b^{+}+S_b^{-}=N.
\]
The intervention accuracy and net change are
\begin{align}
\operatorname{Acc}(b)
&=
\frac{R_b+S_b^{+}}{N},
\label{eq:intervention-accuracy}
\\
\Delta_b
&=
100\frac{R_b-H_b}{N}.
\label{eq:net-change-app}
\end{align}

\paragraph{Exact McNemar test.}
Only the discordant counts $(R_b,H_b)$ enter the paired test. Under the
null of equal marginal error rates,
\[
X\sim\operatorname{Binomial}(R_b+H_b,1/2).
\]
We report the two-sided exact value
{\scriptsize
\begin{equation}
p_{\mathrm{McN}}
=
\min\!\left\{
1,\quad
2\sum_{k=0}^{\min(R_b,H_b)}
\binom{R_b+H_b}{k}2^{-(R_b+H_b)}
\right\}.
\label{eq:exact-mcnemar}
\end{equation}
}

\subsection{Context-Complete Reasoning Track Analysis}
\label{app:intervention-context-sensitivity}

Let $\gamma_i\in\{0,1\}$ be the frozen context-completeness flag and
\[
\mathcal{D}_{\mathrm{cc}}
=
\{i:\gamma_i=1\}.
\]
The Context-Complete Reasoning Track contains
$|\mathcal{D}_{\mathrm{cc}}|=7{,}625$ items. All branch outputs,
public and held-out assignments, and empirical difficulty labels are
inherited from the Full-Coverage Track. We recompute accuracy, transition counts, and McNemar tests on $\mathcal{D}_{\mathrm{cc}}$ without retraining any component or re-banding any item.

\section{RQ2 Diagnostics}
\label{app:rq2-diagnostics}

\subsection{Context-Complete Reasoning Track}
\label{app:rq2-context-complete}

The \emph{Context-Complete Reasoning Track} is a fixed item filter on
the same Full-Coverage Track labels. It excludes context-incomplete items
without recomputing difficulty, so all 7{,}625 retained items keep their
original frozen bands. Table~\ref{tab:rq2-full} reports full matched
outcomes, while Table~\ref{tab:rq2-context-complete} reports this
sensitivity analysis.

\begin{table}[t]
\centering
\scriptsize
\setlength{\tabcolsep}{2.4pt}
\renewcommand{\arraystretch}{1.06}
\begin{tabularx}{\columnwidth}{@{}>{\raggedright\arraybackslash}Xrrrrr@{}}
\toprule
\textbf{Method}
& \textbf{Acc.}
& \textbf{Resc.}
& \textbf{Harm}
& \textbf{$\Delta$}
& \textbf{$p$} \\
\midrule
\multicolumn{6}{@{}l}{\textit{CoT, DeepSeek-R1}} \\
Direct
& 90.64 & -- & -- & -- & -- \\
Function-RAG
& 90.32 & 199 & 223 & $-0.31$ & .263 \\
FunctionGraph-RAG
& 90.87 & 213 & 195 & $+0.24$ & .400 \\
\addlinespace[2pt]
\textit{Oracle ceiling}
& 94.39 & 286 & 0 & $+3.75$ & -- \\
\midrule
\multicolumn{6}{@{}l}{\textit{PoT, GPT-4o}} \\
Direct
& 80.60 & -- & -- & -- & -- \\
Function-RAG
& 79.42 & 200 & 290 & $-1.18$ & $<.001$ \\
FunctionGraph-RAG
& 78.91 & 252 & 381 & $-1.69$ & $<.001$ \\
\quad + Informed verifier
& 80.20 & 233 & 264 & $-0.41$ & .178 \\
\addlinespace[2pt]
\textit{Oracle ceiling}
& 85.26 & 355 & 0 & $+4.66$ & -- \\
\bottomrule
\end{tabularx}
\caption{Matched intervention outcomes on the Context-Complete
Reasoning Track ($N=7{,}625$). Rescue and harm are item transitions
relative to Direct, and $\Delta$ is the net accuracy change in
percentage points. The visually separated oracle ceiling is post hoc,
gold-using, and nondeployable.}
\label{tab:rq2-context-complete}
\end{table}

\subsection{Relevance-Judge and Selector Diagnostics}
\label{app:rq2-judge-stratification}

\begin{table}[t]
\centering
\scriptsize
\setlength{\tabcolsep}{1.8pt}
\renewcommand{\arraystretch}{1.04}
\resizebox{\columnwidth}{!}{%
\begin{tabular}{@{}crrrrrrr@{}}
\toprule
\textbf{$|S_i|$}
& \textbf{$N$}
& \shortstack{\textbf{Function}\\\textbf{RAG}}
& \shortstack{\textbf{FunctionGraph-}\\\textbf{RAG}}
& \textbf{Resc.}
& \textbf{Harm}
& \textbf{$\Delta$}
& \textbf{$p$} \\
\midrule
\multicolumn{8}{@{}l}{\textit{Full-Coverage Track} ($N=10{,}198$)} \\
0 & 7{,}002 & 72.08 & 70.98 & 310 & 387 & $-1.10$ & .004 \\
1 & 2{,}225 & 60.31 & 63.24 & 213 & 148 & $+2.92$ & .001 \\
2 & 479 & 55.53 & 56.99 & 40 & 33 & $+1.46$ & .483 \\
3 & 492 & 58.74 & 59.15 & 31 & 29 & $+0.41$ & .897 \\
\midrule
\multicolumn{8}{@{}l}{\textit{Context-Complete Reasoning Track} ($N=7{,}625$)} \\
0 & 5{,}586 & 80.68 & 79.61 & 184 & 244 & $-1.07$ & .004 \\
1 & 1{,}460 & 75.89 & 77.67 & 97 & 71 & $+1.78$ & .053 \\
2 & 293 & 74.40 & 73.38 & 17 & 20 & $-1.02$ & .743 \\
3 & 286 & 77.97 & 77.27 & 12 & 14 & $-0.70$ & .845 \\
\bottomrule
\end{tabular}
}
\caption{Matched GPT-4o PoT outcomes stratified by $|S_i|$, the number
of functions retained by the Function-RAG LLM relevance judge.
Accuracies compare Function-RAG with FunctionGraph-RAG on the same
items. Rescue and harm
are transitions from Function-RAG, and $\Delta$ is in percentage
points.}
\label{tab:rq2-retained-function-strata}
\end{table}

Retaining zero functions is a harmful regime for FunctionGraph-RAG,
while retaining one function is the clearest favorable regime. The
effects for larger retained sets are smaller and change direction on
the Context-Complete Reasoning Track, so more functions do not yield monotonic
gains.

On the Full-Coverage Track, the zero-function and one-function strata retain
their sign after BH correction, with $q=.0079$ and
$q=.0029$, respectively. The larger retained-function strata are not
significant after correction. This supports an interaction between graph
expansion and the relevance judge's estimate of evidence need, not a
claim that graph topology alone causes the gain.

\paragraph{Exploratory subject effects.}
Credit Risk shows a PoT gain over Function-RAG of $+7.73$ percentage
points (raw $p=.009$, BH $q=.142$). Economics shows a CoT gain over
Direct of $+5.03$ points (raw $p=.052$, $q=.370$), and Quantitative
Methods shows a PoT gain over Function-RAG of $+4.05$ points (raw
$p=.146$, $q=.622$). None survives multiplicity correction, so these
local patterns are hypothesis generating rather than confirmatory.

\subsection{Branch Verification and Oracle Headroom}
\label{app:rq2-verifier-oracle}

On the Full-Coverage Track, the branch verifier changes the GPT-4o PoT
transition counts from 527 rescues and 660 harms for always-on
FunctionGraph-RAG to 484 rescues and 530 harms. It therefore removes 130
harms at the cost of 43 rescues, recovering 0.86 points relative to the
unverified branch. On the Context-Complete Reasoning Track, it removes
117 harms at the cost of 19 rescues, recovering 1.29 points relative to
FunctionGraph-RAG. These counts support a harm-control interpretation and
do not establish a gain over Direct.

The post hoc oracle ceilings exceed Direct by 7.25 points for
DeepSeek-R1 CoT and 7.46 points for GPT-4o PoT on the Full-Coverage Track, and
by 3.75 and 4.66 points on the Context-Complete Reasoning Track. The
frozen branches therefore contain complementary signal, but the oracle
uses gold correctness and remains separate from the deployable RQ3 gate.

\section{RQ3 Diagnostics}
\label{app:rq3-gate}

The adopted Direct-conditioned gate is a strictly held-out selective
branch router. Direct always runs first, after which the gate uses only
the 27 features observable at the Direct stage to decide whether to invoke
FunctionGraph-RAG lazily.

\subsection{Architecture Selection on Public Data}
\label{app:rq3-architecture-selection}

Architecture, features, regularization, and threshold selection use only
five-fold out-of-fold accuracy on the 5{,}110 public items. Held-out
accuracy is computed once after these choices are frozen.

\begin{table}[t]
\centering
\scriptsize
\setlength{\tabcolsep}{2.4pt}
\renewcommand{\arraystretch}{1.06}
\begin{tabular}{@{}lrrrr@{}}
\toprule
\textbf{Variant} & \textbf{Feat.} & \textbf{Branch inv.} & \textbf{OOF} & \textbf{Held.} \\
\midrule
Pairwise selector & 23 cross & 3.00$\times$ & 68.45 & 71.36 \\
Three-class selector & 27 cross & 3.00$\times$ & 66.79 & 70.46 \\
Per-arm binary & 27 cross & 3.00$\times$ & 68.83 & 71.21 \\
\textbf{Direct-conditioned gate} & \textbf{27 direct} & \textbf{1.08$\times$} & \textbf{68.85} & \textbf{71.23} \\
\bottomrule
\end{tabular}
\caption{Candidate routing formulations. Cross-branch selectors inspect
three completed branches before choosing and therefore require three
branch invocations. The adopted gate is selected by public out-of-fold
accuracy and uses only item and Direct-stage features. The invocation
counts are not measurements of tokens, cost, or latency.}
\label{tab:rq3-architecture-selection}
\end{table}

\subsection{Selective Router Training and Leakage Audit}
\label{app:router-details}

\paragraph{Direct-conditioned gate.}
The final router is a gate conditioned on the completed Direct branch.
For every item, Direct runs first and produces output
$o_i^{\mathrm{dir}}$ and Direct metadata $m_i^{\mathrm{dir}}$. The gate
then constructs the 27-dimensional vector
\begin{equation}
z_i
=
\phi\!\left(
x_i,o_i^{\mathrm{dir}},m_i^{\mathrm{dir}}
\right).
\label{eq:router-features}
\end{equation}
It uses this vector to decide whether a lazy deployment would invoke
FunctionGraph-RAG. It is neither a zero-execution pre-gate nor a post
hoc selector that observes both branch outputs.

\paragraph{Feature construction.}
Table~\ref{tab:router-feature-families} groups the 27 fields constructed
by the frozen code. No feature contains the gold answer, a reference
rationale, FunctionGraph-RAG output, retrieval output, or any other
branch information.

\begin{table}[t]
\centering
\scriptsize
\setlength{\tabcolsep}{3.2pt}
\renewcommand{\arraystretch}{1.08}
\begin{tabularx}{\columnwidth}{@{}lc>{\raggedright\arraybackslash}X@{}}
\toprule
\textbf{Feature group} & \textbf{Count} & \textbf{Included signals} \\
\midrule
Direct state & 3
& Unparsed prediction, executor success, and parser success. \\
Direct errors & 4
& Call not allowed, nested function not allowed, missing function
return, and disallowed assignment target. \\
Direct usage & 4
& Input and output token summaries, log latency, and HTTP attempts. \\
Predicted option & 4
& Indicators for Direct predictions A, B, C, and D. \\
Examination and stage & 6
& CFA indicator and indicators for CFA Levels I--III and FRM Parts
I--II. \\
Item form and cues & 6
& Option count, log stem length, numeric option cue, compute cue,
judgment cue, and digit density. \\
\midrule
\textbf{Total} & \textbf{27} & \\
\bottomrule
\end{tabularx}
\caption{The 27 features used by the Direct-conditioned gate. All
features come from the item and the completed Direct branch.}
\label{tab:router-feature-families}
\end{table}

\paragraph{Public-only target construction and sample weighting.}
Let $c_i^{\mathrm{dir}}$ and $c_i^{\mathrm{fg}}$ denote the
correctness indicators of the Direct and FunctionGraph-RAG branches,
respectively. Public gold labels are used to construct a supervised
branch-preference target:

\begin{equation}
t_i
=
\mathbb{I}
\left[
c_i^{\mathrm{fg}}=1
\land
c_i^{\mathrm{dir}}=0
\right],
\qquad
i\in\mathcal{D}_{\mathrm{pub}}.
\label{eq:router-target}
\end{equation}

Thus, FunctionGraph-RAG receives the positive target only when it
strictly improves correctness over Direct. The other three outcomes
favor Direct. In particular, ties are resolved toward Direct when both
branches are correct and when both branches are incorrect. The target
$t_i$ is a branch-preference label rather than an answer-correctness
label. Therefore, $t_i=0$ does not imply that the Direct branch itself
is correct.

All four branch-outcome configurations are retained in training.
We assign each public item a sample weight

\begin{equation}
w_i
=
\begin{cases}
1.00,
& c_i^{\mathrm{dir}}\neq c_i^{\mathrm{fg}},\\[2pt]
\lambda_{\mathrm{tie}},
& c_i^{\mathrm{dir}}=c_i^{\mathrm{fg}},
\end{cases}
\qquad
\lambda_{\mathrm{tie}}=0.15.
\label{eq:router-sample-weight}
\end{equation}

Disagreements receive full weight. The 4{,}483 agreements retain weight
$0.15$ to preserve background supervision while limiting their
influence.

\begin{table}[t]
\centering
\scriptsize
\setlength{\tabcolsep}{3.4pt}
\renewcommand{\arraystretch}{1.08}
\begin{tabular}{@{}cccrr@{}}
\toprule
\textbf{Direct}
& \shortstack{\textbf{FunctionGraph-}\\\textbf{RAG}}
& \textbf{Target $t_i$}
& \textbf{Weight $w_i$}
& \textbf{\# Public} \\
\midrule
Correct
& Correct
& 0
& 0.15
& 3{,}118 (61.0\%) \\

Correct
& Incorrect
& 0
& 1.00
& 352 (6.9\%) \\

Incorrect
& Correct
& 1
& 1.00
& 275 (5.4\%) \\

Incorrect
& Incorrect
& 0
& 0.15
& 1{,}365 (26.7\%) \\
\midrule
\multicolumn{4}{l}{\textbf{Total}}
& \textbf{5{,}110 (100\%)} \\
\bottomrule
\end{tabular}
\caption{Branch-preference targets and sample weights on the public partition. FunctionGraph-RAG is preferred only when it uniquely repairs a Direct error. All public items are retained, while correctness-tie cases are downweighted.
}
\label{tab:router-target-distribution}
\end{table}

Let $h_{\theta}(z_i)$ denote the router's predicted probability of
selecting FunctionGraph-RAG from the inference-visible feature vector
$z_i$. The estimator is fitted through weighted empirical risk
minimization:

{\scriptsize
\begin{equation}
\widehat{\theta}
=
\arg\min_{\theta}
\frac{1}{\sum_{i\in\mathcal{D}_{\mathrm{pub}}}w_i}
\sum_{i\in\mathcal{D}_{\mathrm{pub}}}
w_i\alpha_{t_i}\,
\ell
\left(
h_{\theta}(z_i),
t_i
\right)
+\frac{\lVert\theta\rVert_2^2}{2C},
\label{eq:router-weighted-objective}
\end{equation}
}

where $\alpha_{t_i}$ is the inverse-frequency class weight from
\texttt{class\_weight=balanced}, and $C$ controls L2 regularization.
Model and threshold selection use five-fold
\texttt{StratifiedKFold} cross-validation with shuffling enabled and
seed 202607. The search considers
$C\in\{0.05,0.15,0.5,1.5\}$ and thresholds from 0.40 through 0.94 in
steps of 0.01. This public-only search freezes $C=0.5$ and
$\widehat{\tau}=0.68$. The final classifier is fitted only on the
5{,}110 public items.

Held-out labels, held-out correctness, reference rationales, and
FunctionGraph-RAG outputs enter neither gate features nor fitting,
cross-validation, or threshold selection. Held-out evaluation
decisions were made only after the model and threshold were frozen.
The implementation writes the fitted gate before constructing the held out feature matrix and its whitelist rejects nonpublic identifiers in the fitting path. 
The fitted estimator and threshold are serialized before held-out inference. The training and model-selection paths accept only public identifiers, and a whitelist rejects nonpublic identifiers before feature construction or fitting. Held-out labels, correctness, reference rationales, and FunctionGraph-RAG outputs are never used for gate features, cross-validation, regularization selection, or threshold selection. The held-out partition is evaluated only after the complete gate configuration has been frozen.

\subsection{Strict Held-Out Gate Results}
\label{app:rq3-heldout-results}

\paragraph{Frozen evaluation and lazy-deployment interpretation.}
The reported evaluation applies the frozen gate to precomputed frozen
Direct and FunctionGraph-RAG predictions. The gate observes only item
and Direct features when it computes
\begin{equation}
p_i
=
h_{\widehat{\theta}}(z_i),
\qquad
r_i
=
\mathbb{I}[p_i\geq 0.68].
\label{eq:router-probability}
\end{equation}
The evaluated selection rule is
\begin{equation}
\hat{y}^{\,\mathrm{gate}}_i =
\begin{cases}
\hat{y}^{\,\mathrm{dir}}_i,
& r_i=0,\\
\hat{y}^{\,\mathrm{fg}}_i,
& r_i=1.
\end{cases}
\label{eq:router-decision-app}
\end{equation}

The trigger rate and implied branch invocation count under a lazy
deployment are
\begin{equation}
\begin{aligned}
\operatorname{TriggerRate}
&=
\frac{1}{N}
\sum_{i=1}^{N} r_i,
\\
\operatorname{BranchInvocations}
&=
1+\operatorname{TriggerRate}.
\end{aligned}
\label{eq:selection-rate}
\end{equation}
Under that deployment, Direct contributes one invocation and each
trigger adds one FunctionGraph-RAG invocation. The released evaluation
did not execute this policy lazily. The implied count is not measured
tokens, latency, money, or energy.

Public gold labels affect only the training targets $t_i$ and sample
weights $w_i$. Neither quantity is available to the router during
held-out inference. The held-out decision $r_i$ depends only on the
frozen estimator, the frozen threshold, and the inference-visible
feature vector $z_i$.

\paragraph{Explored selector comparison.}
Table~\ref{tab:router-ablation} separates the adopted gate from three
explored selectors. The explored selectors observe three completed
branches. None is an oracle.

\begin{table}[t]
\centering
\scriptsize
\setlength{\tabcolsep}{2.0pt}
\renewcommand{\arraystretch}{1.08}
\begin{tabularx}{\columnwidth}{@{}>{\raggedright\arraybackslash}Xrrrr@{}}
\toprule
\textbf{Selector}
& \textbf{Acc.}
& \textbf{$\Delta$}
& \textbf{$p_{\mathrm{McN}}$}
& \textbf{Invocations} \\
\midrule
Legacy pairwise (23 features, $\tau=.70$) & 71.36 & +0.53 & 0.107 & 3 \\
Multiclass argmax (27 features) & 70.46 & $-0.37$ & 0.475 & 3 \\
Per-condition binaries (27 features) & 71.21 & +0.37 & 0.252 & 3 \\
Direct-conditioned gate (27 features) & 71.23 & +0.3931 & 0.0446 & $1.079\times$ \\
\bottomrule
\end{tabularx}
\caption{Held-out routing variants. The legacy pairwise selector
is post hoc and uses a three-branch budget. Accuracy and $\Delta$ are
percentage points relative to Direct. The final row is the adopted
gate.}
\label{tab:router-ablation}
\end{table}

\subsection{Oracle Selection Ceiling}
\label{app:oracle-ceiling}

The oracle ceiling is computed post hoc only after every branch output is frozen.
For a branch set $\mathcal{B}$, its item-level correctness is
\begin{equation}
c_i^{\mathrm{oracle}}
=
\max_{b\in\mathcal{B}}c_i^b,
\label{eq:oracle-correctness}
\end{equation}
and
\begin{equation}
\operatorname{Acc}_{\mathrm{oracle}}
=
\frac{1}{N}
\sum_{i=1}^{N}
c_i^{\mathrm{oracle}}.
\label{eq:oracle-accuracy}
\end{equation}

This quantity uses gold correctness to select a branch and is therefore
nondeployable. It is a post hoc gold-using union upper bound, not the
learned router, and does not contribute to router fitting, threshold
selection, or held-out prediction.

\subsection{Bounded Agentic Probes}
\label{app:post-execution-verifier}
\label{app:bounded-agentic-probes}

We evaluate three exploratory probes, each with a single backbone and a fixed budget. They are diagnostics rather than complete agentic frameworks. Endpoint accuracy uses every item in the listed population as its denominator. Selective accuracy instead conditions on the items accepted by the abstention detector and therefore does not measure end-to-end accuracy.

\begin{table*}[t]
\centering
\small
\setlength{\tabcolsep}{4pt}
\begin{tabularx}{\textwidth}{@{}>{\raggedright\arraybackslash}p{0.17\textwidth}>{\raggedright\arraybackslash}p{0.23\textwidth}>{\raggedright\arraybackslash}X>{\raggedright\arraybackslash}p{0.26\textwidth}@{}}
\toprule
Probe & Population & Outcome & Statistical result \\
\midrule
Post-execution verification
& 369 Context-Complete Hard items with a produced program
& Flags 13.6\%; revises 12.7\%; 11 rescues and 10 harms
& Endpoint $\Delta=+0.27$ points; exact paired $p=1.000$ \\
Abstention detection
& 1{,}199 items
& Accepted-item accuracy: 57.6\% baseline, 85.9\% selective
& Matched-random selective accuracy: 57.6\%; answerable-item FPR: 53.3\% \\
Plan then solve
& 300 items
& Endpoint $\Delta=-3.33$ points; Easy $+26.9$; Hard $-25.6$
& Paired $p=.353$; below chance on the subset failed by every static condition \\
\bottomrule
\end{tabularx}
\caption{Bounded agentic probes. Endpoint changes retain the full listed population as the denominator. Accepted-item accuracy is selective and conditions on coverage.}
\label{tab:bounded-agentic-probes}
\end{table*}

For post-execution verification, a flag marks an answer as implausible after checking unit, sign, magnitude, and answer-option consistency; rescues and harms are incorrect-to-correct and correct-to-incorrect transitions. The flag and revision rates use all 369 produced-program items as their denominator. For abstention, the false-positive rate is the fraction of answerable items incorrectly rejected. For plan then solve, the Easy--Hard contrast suggests that explicit planning can faithfully execute an initially wrong approach; it does not identify a causal mechanism. Across the three probes, none yields a reliable end-to-end gain. Their pattern is consistent with shared misconceptions, not proof of them.

\subsubsection{Post-Execution Verification}
\label{app:postexecution-verification}

The diagnostic covers 369 Context-Complete Hard items for which the PoT pipeline produces an executable program. After execution, the same backbone receives the program, numerical result, and selected option, and checks sign, units, magnitude, and answer-option consistency. The verifier flags 50 items (13.6\%) and changes 47 answers (12.7\%). It yields 11 rescues and 10 harms, for $+0.27$ points, a two-sided exact McNemar value of $p=1.000$, and a bootstrap 95\% confidence interval of $[-2.17,+2.71]$ points. Without independent evidence, same-backbone verification has limited leverage against errors shared with the solver.

\subsubsection{Answerability Detection and Selective Abstention}
\label{app:answerability-abstention}

We sample 1{,}199 items stratified by context completeness and empirical
difficulty. The sample is approximately balanced between locally
answerable and context-dependent records and is therefore not
representative of the benchmark's natural 74.8/25.2 prevalence.

\begin{table}[t]
\centering
\scriptsize
\setlength{\tabcolsep}{3.5pt}
\renewcommand{\arraystretch}{1.08}
\begin{tabular}{@{}lrr@{}}
\toprule
& \textbf{Pred. answerable} & \textbf{Pred. unanswerable} \\
\midrule
Expert answerable & 280 & 320 \\
Expert unanswerable & 4 & 595 \\
\bottomrule
\end{tabular}
\caption{Answerability-detection confusion matrix on the stratified
audit. ``Unanswerable'' is the positive class.}
\label{tab:answerability-confusion}
\end{table}

The detector reaches 72.98\% accuracy, 65.03\% precision, 99.33\%
recall, and 78.60 F1 for the unanswerable class. If the model answers only items predicted answerable, selective accuracy rises from 57.55\% at full coverage to 85.92\% at 23.7\% coverage. Matched-rate random abstention averages 57.56\%, and none of 4{,}000 simulations reaches the detector's selective accuracy. The false-positive rate on expert-answerable items is 53.3\%, so this is a risk signal rather than a calibrated deployment policy.

\subsubsection{Plan-Then-Solve}
\label{app:plan-then-solve}

The paired sample contains 300 questions. Half are missed by all three static conditions and half are answered correctly by Direct, with the two groups matched by empirical difficulty. Direct accuracy is 50.00\% by construction. Plan-then-solve reaches 46.67\%, with 42 rescues and 52 harms. The net change is $-3.33$ points, the two-sided exact McNemar test gives $p=.3533$, and the bootstrap 95\% confidence interval is $[-9.67,+3.00]$ points.

\begin{table}[t]
\centering
\small
\setlength{\tabcolsep}{5.0pt}
\renewcommand{\arraystretch}{1.08}
\begin{tabular}{@{}lrrr@{}}
\toprule
\textbf{Band} & \textbf{Direct} & \textbf{Plan-then-solve} & \textbf{$\Delta$} \\
\midrule
Easy & 50.00 & 76.92 & $+26.92$ \\
Medium & 50.00 & 42.86 & $-7.14$ \\
Hard & 50.00 & 24.39 & $-25.61$ \\
\bottomrule
\end{tabular}
\caption{Descriptive band-wise results in the difficulty-matched
plan-then-solve sample. Direct is fixed at 50\% within each band by
construction.}
\label{tab:plan-by-band}
\end{table}

\subsubsection{Joint Interpretation}
\label{app:agentic-probe-interpretation}

The two deliberation probes do not improve paired accuracy. Selective abstention carries useful information but obtains it by rejecting many answerable items. These findings do not establish that agentic financial reasoning is ineffective. They isolate a narrower limitation: reusing the same backbone over the same evidence is insufficient when the initial error reflects a stable misconception.

\subsection{Scope Relative to Agentic Reasoning}
\label{app:agentic-scope}

Under a lazy deployment, the gate would make one decision after Direct returns and before any optional FunctionGraph-RAG invocation. It does not make an earlier need decision or learn a trajectory of retrieval and tool actions. A future multi-component agentic framework could instead maintain a state $h_t$ and choose actions
\[
\begin{aligned}
a_t\in\{&\textsc{Retrieve},\textsc{ExpandGraph},\textsc{Execute},\\
&\textsc{Verify},\textsc{Revise},\textsc{Stop}\}.
\end{aligned}
\]
This formulation would support earlier decisions about additional computation, query reformulation, multiple tool calls, execution feedback, revision, and dynamic stopping. Evaluating such policies under bounded budgets and without exposing held-out questions is left to future work.

\section{Reproducibility Details}
\label{app:reproducibility}

\subsection{Model Access and Hardware}
\label{app:hardware}

The fixed 17-model panel contains seven open-weight models deployed locally and ten models accessed through official or provider-hosted APIs. We used local deployment whenever the released checkpoint fit the available hardware. DeepSeek-R1 was evaluated through an API despite its open license because its 671B mixture-of-experts checkpoint could not be served on the same cluster under the benchmark context and completion budgets.

All local models were served with vLLM on NVIDIA H100 accelerators. The five finance-specialized models used bfloat16 weights. DianJin-R1-32B required two H100 GPUs with tensor parallelism of two. Fin-O1-14B, ODA-Fin-RL-8B, Hawkish-8B, and Fin-R1-7B each ran on one H100. The two GPT-OSS models used their released MXFP4 weights with bfloat16 computation. GPT-OSS-20B ran on one H100 with batch size four, while GPT-OSS-120B used batch size one to preserve key-value-cache capacity. Table~\ref{tab:model-access-config} reports the exact checkpoint revisions and serving configuration.

\begin{table*}[t]
\centering
\small
\setlength{\tabcolsep}{4pt}
\renewcommand{\arraystretch}{1.08}

\begin{tabularx}{0.88\textwidth}{@{}l >{\raggedright\arraybackslash}X l@{}}
\toprule
\multicolumn{3}{@{}l}{\textit{Proprietary, API-served}} \\
\midrule
\textbf{Model} & \textbf{Model identifier} & \textbf{Reasoning setting} \\
\midrule
Claude-Sonnet-5 & \texttt{anthropic/claude-sonnet-5}     & provider default \\
\addlinespace[2pt]
DeepSeek-V4-Pro & \texttt{deepseek/deepseek-v4-pro}      & high \\
DeepSeek-R1     & \texttt{deepseek/deepseek-reasoner}    & provider default \\
\addlinespace[2pt]
Gemini-3.1-Pro  & \texttt{google/gemini-3.1-pro-preview} & low \\
\addlinespace[2pt]
GPT-5.6-Sol     & \texttt{openai/gpt-5.6-sol}            & provider default \\
GPT-5.6-Terra   & \texttt{openai/gpt-5.6-terra}          & provider default \\
GPT-5.6-Luna    & \texttt{openai/gpt-5.6-luna}           & provider default \\
GPT-5.5         & \texttt{openai/gpt-5.5}                & none \\
GPT-4o          & \texttt{openai/gpt-4o-2024-11-20}      & provider default \\
\addlinespace[2pt]
Qwen3.7-Max     & \texttt{qwen/qwen3.7-max}              & provider default \\
\bottomrule
\end{tabularx}

\vspace{0.8em}

\begin{tabularx}{0.88\textwidth}{@{}l >{\raggedright\arraybackslash}X l c c c@{}}
\toprule
\multicolumn{6}{@{}l}{\textit{Open-weight, locally deployed}} \\
\midrule
\textbf{Model} & \textbf{Hugging Face identifier} & \textbf{Precision} & \textbf{TP} & \textbf{GPUs} & \textbf{Batch} \\
\midrule
\multicolumn{6}{@{}l}{\hspace{2pt}\textit{General-purpose reasoning}} \\
GPT-OSS-120B   & \texttt{openai/gpt-oss-120b}              & MXFP4/BF16 & 1 & 1 & 1 \\
GPT-OSS-20B    & \texttt{openai/gpt-oss-20b}               & MXFP4/BF16 & 1 & 1 & 4 \\
\midrule
\multicolumn{6}{@{}l}{\hspace{2pt}\textit{Finance-specialized}} \\
DianJin-R1-32B & \texttt{DianJin/DianJin-R1-32B}           & BF16       & 2 & 2 & 2 \\
Fin-O1-14B     & \texttt{TheFinAI/Fin-o1-14B}              & BF16       & 1 & 1 & 4 \\
ODA-Fin-RL-8B  & \texttt{OpenDataArena/ODA-Fin-RL-8B}     & BF16       & 1 & 1 & 4 \\
Hawkish-8B     & \texttt{mukaj/Llama-3.1-Hawkish-8B}       & BF16       & 1 & 1 & 4 \\
Fin-R1-7B      & \texttt{SUFE-AIFLM-Lab/Fin-R1}           & BF16       & 1 & 1 & 4 \\
\bottomrule
\end{tabularx}

\caption{Model access and serving configurations. Proprietary parameter counts are omitted because they are not publicly disclosed. Reasoning settings use provider defaults unless otherwise shown. All locally deployed model runs use a $32{,}768$-token context window and an $8{,}192$-token completion budget on NVIDIA H100 accelerators.}
\label{tab:model-access-config}
\end{table*}

\subsection{Software Environment and Run Integrity}

All local evaluations used Python~3.11, vLLM~0.10.0, Transformers~4.53.2, Hugging Face Hub~0.33.4, tokenizers~0.21.2, and safetensors~0.5.3. Each run produced a configuration hash covering the model and tokenizer revisions, prompt template, decoding parameters, numeric precision, context length, completion budget, batch size, tensor-parallel degree, random seed, and core software versions. A resumed job was accepted only when this hash matched the original run. This prevents a partially completed evaluation from being silently resumed under a different model revision or inference environment.

The benchmark denominator is fixed at 10{,}198 items for every model. Output files are normalized to the canonical item order, and duplicate records are resolved deterministically by retaining the final record for an item identifier. The normalization step then verifies that the resulting identifier set matches the full benchmark exactly.

\subsection{Unified Inference Protocol}

We use one direct multiple-choice protocol across the fixed 17-model panel. Local runs use temperature~$0$, top-$p$~$1$, a 32{,}768-token context window, and an 8{,}192-token completion budget. This common setting is an evaluation choice rather than a reproduction of each model card's recommended generation configuration. In particular, Fin-O1-14B and ODA-Fin-RL-8B recommend temperature~$0.6$ and top-$p$~$0.95$, while Fin-R1-7B recommends temperature~$0.7$ and top-$p$~$0.8$. We hold decoding fixed to isolate model differences from sampling differences.

For API endpoints, we requested the closest available deterministic configuration and recorded any provider-side exception. Some reasoning endpoints do not expose temperature, top-$p$, or a reproducible seed. Reasoning-budget controls were set explicitly when available. In the frozen panel, GPT-5.5 used \texttt{none}, Gemini-3.1-Pro used \texttt{low}, and DeepSeek-V4-Pro used \texttt{high}. These settings are reported with the corresponding leaderboard entries. An item whose prompt and completion budget exceeded the supported context window was recorded as an evaluation failure rather than silently truncated from the input.

\subsection{Prompt and Output Schema}
\label{app:prompts}

Local checkpoints receive the benchmark prompt through their released chat template using \texttt{apply\_chat\_template}, with the generation prompt enabled. The resulting token sequence is passed directly to vLLM. API models receive the same user text through their standard endpoints. We do not add a shared model message because several finance-specialized checkpoints were tuned with a user-only instruction format.

The prompt contains only the question and its answer options. Gold labels and source-provided rationales are never included in model input. CFA items contain three options and FRM items contain four. We preserve the option identifiers supplied by each item rather than relabeling the choices during evaluation.

\begin{figure}[H]
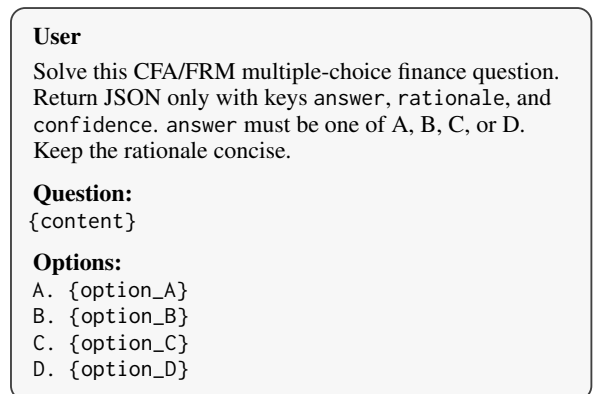

\centering
\begin{promptbox}
\raggedright
\textbf{User}

\smallskip
Solve this CFA/FRM multiple-choice finance question. Return JSON only
with keys \texttt{answer}, \texttt{rationale}, and
\texttt{confidence}. \texttt{answer} must be one of A, B, C, or D.
Keep the rationale concise.

\medskip
\textbf{Question:}

\texttt{\{content\}}

\medskip
\textbf{Options:}

\texttt{A. \{option\_A\}}\\
\texttt{B. \{option\_B\}}\\
\texttt{C. \{option\_C\}}\\
\texttt{D. \{option\_D\}}
\end{promptbox}
\caption{The common user prompt used across models and evaluation
conditions.}
\label{fig:prompt-mcq}
\end{figure}

\end{document}